%% file: main.tex
\documentclass[10pt,twocolumn,letterpaper]{article}

 \usepackage[pagenumbers]{style} %

\input{preamble}

\input{math_commands.tex}

\usepackage{xr}
\definecolor{styleblue}{rgb}{0.21,0.49,0.74}
\usepackage[pagebackref,breaklinks,colorlinks,allcolors=styleblue]{hyperref}

\def\paperID{} %
\def\confName{}
\def\confYear{}

\title{3D Shape Tokenization via Latent Flow Matching}

\author{
	Jen-Hao Rick Chang,
	Yuyang Wang,
	Miguel Angel Bautista Martin \\
	Jiatao Gu,
	Xiaoming Zhao,
	Josh Susskind,
	Oncel Tuzel \\
	Apple \\
	{\small \url{https://machinelearning.apple.com/research/3d-shape-tokenization}}
}

\begin{document}

\twocolumn[{%
	\renewcommand\twocolumn[1][]{#1}%
	\maketitle
	 \vspace{-5mm}
	\centering
	\includegraphics[width=\linewidth]{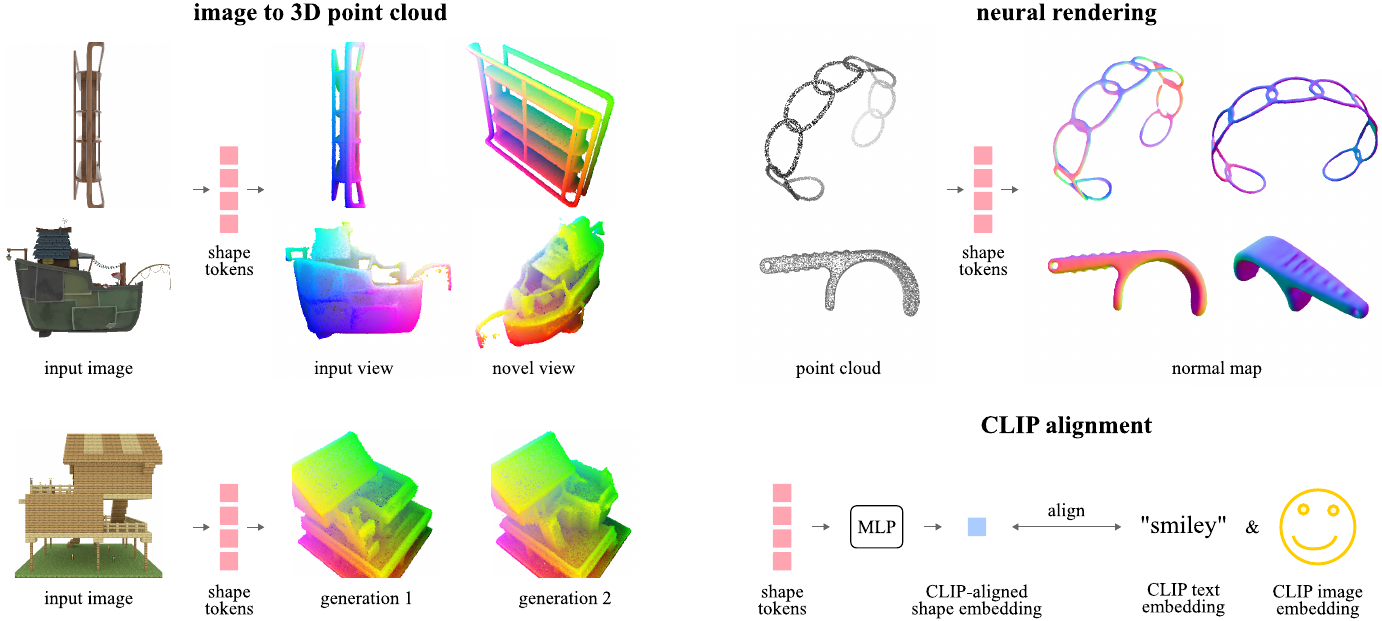}
	\vspace{-6mm}
	\captionof{figure}{
		\textit{\Ours} can be readily used as input / target to machine learning models in various applications, including single-image-to-3D (left), neural rendering of normal maps (top right) and 3D-CLIP alignment (bottom right).
		Mesh credits~\cite{whaarquitectos_librero_repisas, rodiergabrielle_early_morning, madexc_domik_house, bennettgraham_bracelet, andresblancof_gohome1}.
		\vspace{2em}}
	\label{fig: teaser}
}]

\input{sec/0_abstract_v2}

\input{sec/1_intro_v2}

\input{sec/2_related}

\input{sec/3_method}

\input{sec/4_exp}

\input{sec/5_discussion}

{
    \small
    \bibliographystyle{ieeenat_fullname}
    \bibliography{main}
}

\clearpage
\appendix

\maketitlesupplementary

\input{sec/99_appendix}

\end{document}

%% file: preamble.tex
\usepackage{enumitem}
\usepackage{subcaption}
\usepackage{comment}
\usepackage[utf8]{inputenc}

\usepackage{mathtools}

\usepackage{xspace}
\makeatletter
\DeclareRobustCommand\onedot{\futurelet\@let@token\@onedot}
\def\@onedot{\ifx\@let@token.\else.\null\fi\xspace}

\renewcommand{\emph}[1]{\textit{{#1}}}
\def\eg{\emph{e.g}\onedot} 
\def\ie{\emph{i.e}\onedot} 
 
\def\etc{\emph{etc}\onedot} 
\def\wrt{w.r.t\onedot} 

\def\iid{i.i.d\onedot}

\makeatother

\usepackage{sistyle}
\SIthousandsep{,}

\usepackage{upgreek}
\usepackage{makecell}

\input{definitions}

\usepackage{physics}
\usepackage[export]{adjustbox}%

\usepackage{array}
\newcolumntype{C}[1]{>{\centering\arraybackslash}m{#1}}  %

\usepackage{multirow}

\definecolor{orange}{rgb}{0.8, 0.6, 0.2}
\definecolor{lightgrey}{rgb}{0.7, 0.7, 0.7}

\newcommand{\blue}[1]{\textcolor{blue}{#1}}

\newcommand{\TOPIC}[1]{\textcolor{lightgrey}{\bf [TOPIC: #1]}}
\renewcommand{\TOPIC}[1]{} %

\definecolor{rtwocolor}{RGB}{50,50,200}
\definecolor{ronecolor}{RGB}{200,50,50}
\definecolor{rthreecolor}{RGB}{50,200,50}

\newcommand{\cmark}{{\ding{51}}}%

\usepackage{placeins}
\usepackage{xr}

\usepackage{colortbl}
\newcommand{\graymidrule}{\arrayrulecolor{gray!50} \midrule \arrayrulecolor{black}}

\renewcommand{\paragraph}[1]{\vspace{0.5em}\noindent\textbf{#1}}

\newcommand{\Ours}{Shape Tokens\xspace}
\newcommand{\OursABV}{ST\xspace}

\newcommand{\xz}[1]{{{#1}}}

\usepackage{pifont}
\usepackage{rotating}
\usepackage{threeparttable}

\newcommand{\Suppweb}{On the website}
\newcommand{\suppweb}{on the website}
\newcommand{\indexhtml}{on the website}

%% file: definitions.tex
\def\1n{\mathbf{1}_n}
\def\0{\mathbf{0}}
\def\1{\mathbf{1}}

\def\R{{\mathbb R}}

\def\e{{\bf e}}

\DeclareMathOperator*{\bbE}{\mathbb{E}}

\newcommand{\cX}{\mathcal{X}}

\newcommand{\cN}{\mathcal{N}}

\newcommand{\cS}{\mathcal{S}}
\newcommand{\cY}{\mathcal{Y}}
\newcommand{\cZ}{\mathcal{Z}}

\newcommand{\bs}{s}

\newcommand{\ms}{ms\xspace}

%% file: math_commands.tex
\usepackage{amsmath,amsfonts,bm}

\def\eqref#1{equation~\ref{#1}}

\def\1{\bm{1}}

\DeclareMathAlphabet{\mathsfit}{\encodingdefault}{\sfdefault}{m}{sl}
\SetMathAlphabet{\mathsfit}{bold}{\encodingdefault}{\sfdefault}{bx}{n}

%% file: sec/0_abstract_v2.tex
\begin{abstract}

We introduce a latent 3D representation that models 3D surfaces as probability density functions in 3D, \ie, $p(x,y,z)$, with flow-matching.
Our representation is specifically designed for consumption by machine learning models, offering continuity and compactness by construction while requiring only point clouds and minimal data preprocessing.
Despite being a data-driven method, our use of flow matching in the 3D space enables interesting geometry properties, including the capabilities to perform zero-shot estimation of surface normal and deformation field.
We evaluate with several machine learning tasks, including 3D-CLIP, unconditional generative models,  single-image conditioned generative model, and intersection-point estimation.
Across all experiments, our models achieve competitive performance to existing baselines, while requiring less preprocessing and auxiliary information from training data.

\end{abstract}

%% file: sec/1_intro_v2.tex
\section{Introduction}
\label{sec:intro}

The choice of 3D representation is usually determined by the downstream task of interest.
For graphics applications, one may choose a mesh or a 3D Gaussian representation \citep{kerbl3Dgaussians}.
For scientific or physics simulation settings, continuous representations like distance fields might be able to encode fine-grained information \citep{bridson2005simulation, guendelman2003nonconvex}.
For training machine learning (ML) models, we are often looking for a 3D representation that is 1) continuous and 2) compact, due to computation and memory constraints; we would also prefer one that 3) requires minimal preprocessing, to allow learning from large amount of data.
While representations like voxels \citep{deng2021voxel, kim20133voxel, liu2019pointvoxel}, meshes \citep{nash2020polygen, siddiqui2024meshgpt}, point clouds \citep{li2018point, qi2017pointnet, yang2019pointflow,sun2018pointgrow, ddpm_pointcloud, yang2019pointflow, zhang2025gs}, (un-)signed distance fields \citep{chibane2020neural, liu2023neudf, autosdf2022}, radiance/occupancy fields \citep{mildenhall2020nerf, mescheder2019occupancy, peng2020convoccnet, neural_fields}, Gaussian splats \citep{kerbl3Dgaussians, lassner2021pulsar}, and latent representations \citep{zhang2024clay,zhao2023michelangelo,zhang20233dshape2vecset} have been used in learning systems, these representations often do no meet all of the three criteria.  %

In the paper, we are interested in designing a 3D representation to be consumed by ML models. 
We adopt the perspective of \citet{ddpm_pointcloud} and \citet{yang2019pointflow} that treats a 3D shape $\cS$ as a probability density function in 3D space whose probability density concentrates on the (external and internal) surfaces (\ie, $p(xyz) = P\{xyz \, {\in} \, \cS\}$).
Under this construction, sampling $p(xyz)$ produces points on the 3D surface, and a point cloud contains many samples of the 3D distribution, allowing one to fit $p(xyz)$.
We show that by utilizing flow matching \cite{lipman2023flow, albergo2023building} to learn a shared latent space of 3D shapes where each latent vector $\bs$ parameterizes $p(xyz | \bs)$, the resulted latent representation satisfies the aforementioned three criterion by construction. 
We call the latent 3D representation \textit{\Ours (ST)}, and it offers several desirable properties:\footnote{Note that "token" commonly denotes a discrete set of symbols in language models. \Ours refer to a set of real-valued vectors.}
\begin{enumerate}[leftmargin=*]
	\item \OursABV are continuous and compact by construction.  We intentionally designed the representation to be consumed by ML models, utilizing a low-dimensional continuous latent space. Specifically, \OursABV represent diverse shapes using just 1,024 continuous vectors of $16$ dimensions, making \OursABV easy to integrate into ML applications.
	\item Our approach makes minimal assumptions about 3D shapes. We only need independent and identically distributed (\iid) samples of $p(xyz)$, \ie, point clouds sampled from 3D surfaces. Point cloud can represent most 3D shapes, including non-watertight, self-intersecting, and even partial ones, and is the output of most depth sensors.  
    This differentiates our method from many existing latent representations that depend on signed distance functions and occupancy field \cite{zhang2024clay, zhao2023michelangelo} (which need water-tight shapes).  The minimal requirements of \OursABV significantly simplifies our training pipeline and enables us to scale our training set, particularly valuable since most meshes in large-scale datasets like Objaverse~\cite{objaverse} are not watertight and typically difficult to process. 
	\item Interesting, even though our method makes minimal assumption about 3D data and is purely data-driven, our flow-matching approach in 3D space provides geometric capabilities like zero-shot surface normal estimation and deformation between shapes (see \Cref{sec: analysis}).
\end{enumerate}

We empirically investigate the effectiveness of our representation on a range of downstream ML applications (\Cref{fig: teaser}).
First, we align \OursABV to the pretrained image and text CLIP embeddings~\cite{liu2024openshape} by learning a Multi-Layer Perceptron (MLP) adapter, evaluated with zero-shot text classification of 3D shapes (see \Cref{sec: exp clip}).
Second, we tackle 3D generation problems by learning an unconditional flow-matching model on ShapeNet~\cite{shapenet2015} and an image-conditioned flow-matching model on Objaverse (see \Cref{sec: exp lfm}).
Third, we train a neural network to estimate ray-surface interaction, treating the geometry problem as a learning problem. The model takes \OursABV and a ray as input and estimates the intersection point and its normal (see \Cref{sec: exp rendering}).
In all these tasks, we achieve competitive performance as baselines designed for the specific tasks.
We will release pretrained shape tokenizers, image-conditioned latent flow-matching models, the 3D-CLIP model, our data rendering pipeline, and our training code.

%% file: sec/2_related.tex
\section{Related work}
\label{sec: related}

The field of 3D representation, generation, and classification is vast.
We focus on discussing literature most relevant to our work ---latent 3D representations.
For an overview of 3D representations as a whole, we refer reader to~\cite{adv_in_neural_rendering}.

\input{sec/2_1_3d_representation}

\input{sec/2_2_flow_matching}

%% file: sec/2_1_3d_representation.tex
\label{sec: related work 3d}

Latent 3D representations can be categorized by the modeled entities.
We provide a short discussion here.  In appendix (\Cref{table: related works}), we provide more detailed discussions.
3DShape2VecSet~\cite{zhang20233dshape2vecset}, Michelangelo~\cite{zhao2024michelangelo}, Direct3D~\cite{wu2024direct3d}, and Clay~\cite{zhang2024clay} encode surfaces by learning to reconstruct occupancy fields.
While these methods share similar network architecture (\eg, transformer) as ours, training these models requires watertight meshes, which often need extensive preprocessing and filtering to create.
For example, the closing-mesh preprocessing required to acquire watertight meshes often results in loss of quality. %
In comparison, our method can directly train on point clouds sampled from the original shapes without any modification. 
LION~\cite{vahdat2022lion} learns a latent representation of point sets of a fixed cardinality, \ie, they learn the joint distribution of a fixed number of points, \ie, $p(x_1, \dots, x_k)$.
Despite its permutation-invariance, it is a much higher dimensional function (\ie, $3k$) compared to our 3-dimensional distribution, which allows us to use compact latent while achieving similar reconstruction quality.
Additionally, modeling shapes as 3-dimensional distributions with flow matching also enables zero-shot surface normal estimation.
As mentioned in \Cref{sec:intro}, DDPM-points~\citep{luo2021diffusion} and PointFlow~\citep{yang2019pointflow} also consider 3D shapes as 3D probability density functions and use a generative model (diffusion model and continuous normalizing flow, respectively) to model the distributions.
These methods are trained on ShapeNet dataset.
In comparison, our use of flow matching simplifies training and enables us to scale up from ShapeNet to Objaverse.
Moreover, we demonstrate connections between the predicted velocity field and geometric properties like surface normal and deformation fields, and we show that the learned 3D representation is useful for ML applications beyond generative modeling, \eg, zero-shot text classification.

Recently, there emerge many concurrent works that utilize latent 3D representations.
TRELLIS~\cite{xiang2024structured} utilizes sparse voxel grids and multiview features; Dora~\cite{chen2024dora} and Pandora3D~\cite{yang2025pandora3d} propose new point sampling / attention mechanisms and model occupancy field;  TripoSG~\cite{li2025triposg} and Hunyuan3D 2.0~\cite{zhao2025hunyuan3d} model signed distance functions.
Their representations model occupancy or multiview images and utilize total latent dimension much higher than ours, \eg, by an order of magnitude.  They also need a large number of multiview images or extensive data preprocessing to get watertight meshes.
In comparison, our goal is to study the use of flow matching in learning tokenization of 3D shapes and we target generic ML applications like aligning with pretrained CLIP---motivating us to choose a low-dimensional latent space that are more efficient for a wider range of ML applications.
Despite our lower dimensional latent, we show that our method is able to achieve comparable geometry quality as TRELLIS that uses more training information and a higher total latent dimension.

A concurrent work by \citet{zhang2024geometry} proposes to fit individual 3D shape with a diffusion model separately. In comparison, our goal is to learn a general tokenizer that can be applied to wide range of shapes without per-shape optimization.  
\citet{chen2025diffusion} demonstrate advantages in learning image tokenization with diffusion models. While also using diffusion to learn a latent space, they model the distribution of an entire image (thus the distribution dimension is $h{\times}w{\times}3$), \ie, one sample of the distribution  produces an entire image. In contrast, our probability density function (of dimension 3) is the 3D shape itself---each sample is a single point on the shape. This difference allows us to additionally connect the latent representation with geometric properties like surface normal.

%% file: sec/2_2_flow_matching.tex
\section{Preliminary}
\label{sec: related flow matching}

We provide a preliminary overview of flow matching. 
Flow matching generative models~\cite{lipman2023flow, sit} learn to reverse a time-dependent process that turns data samples $x  \sim p(x)$ into noise $\epsilon \sim e(\epsilon)$:  $x_t = \alpha_t x + \sigma_t \epsilon$, 
where $t \in [0, 1]$, $x$ and $\epsilon \, {\in} \, \R^d$, $\alpha_t$ is an increasing function of $t$ and $\sigma_t$ is a decreasing function of $t$, and $p_0(x) \equiv e(x)$ is the distribution of noise and $p_1(x) \approx p(x)$ is the data distribution.  
The marginal probability distribution $p_t(x)$ is equivalent to the distribution of the probability flow Ordinary Differential Equation (ODE) of the following velocity field \citep{sit}: 
\begin{equation}
    \label{eq:prob_flow}
    \dot{x_t} = \dv{x_t}{t} = v_\theta(x; t),
\end{equation}
where $v_\theta(x; t)$ can be learned by minimizing the loss
\begin{equation}
\mathcal{L}(\theta) = \int_{0}^{1} \bbE\left[ \| v_\theta(x_t; t) - \dot{\alpha_t} x_1 - \dot{\sigma_t} \epsilon \|^2 \right] \dd{t}.
\label{eq: flow matching loss}
\end{equation}
In practice, the integrations of time and the expectation are approximated by Monte Carlo methods, allowing simple implementations.
Under this formulation, samples of $p_1(x)$ are generated by integrating the ODE (\cref{eq:prob_flow}) from $t \,{=}\, 0$ to $t \,{=}\, 1$. 
Note that the formulation allows flexible choices of $e(\epsilon)$, $\alpha_t$, and $\sigma_t$, which we utilize in our paper, for more details we refer readers to \cite{lipman2023flow, sit}.

%% file: sec/3_method.tex
\section{Method}
\label{sec: method}

\begin{figure*}[t]
	\centering
	\includegraphics[width=\linewidth]{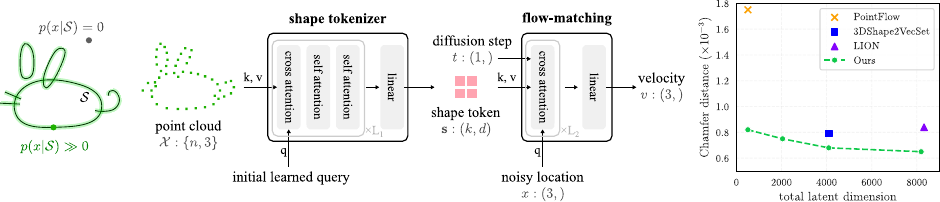}
	\vspace{-7mm}
	\caption{Overview of our architecture.  (Left) We model a 3D shape as a probability density function that is concentrated on the surface, forming a delta function in 3D.  
    (Center) Our tokenizer uses flow matching to learn $p(xyz | \bs)$ and the shape tokenizer.
    (Right) The figure shows the total latent dimension and reconstruction error of various methods trained on ShapeNet dataset. Our tokenizers achieve better trade-off between compactness and reconstruction quality than baselines.
    }
	\label{fig: architecture}
\end{figure*}

We consider a 3D shape $\cS$ as a probability density function $p_{\cS}(x): \R^3 \rightarrow [0, \infty)$, where $x \in \R^3$ is a 3D location (\ie, xyz).  %
A set of \iid samples of $p_{\cS}(x)$ creates a point cloud, $\cX = \left\{ x_1, \dots, x_n \right\}$.
Our goal is to fit $p_{\cS}(x)$ with a conditional flow matching model $v_\theta(x; \bs, t): \R^3 \rightarrow \R^3$, where $t \in [0, 1]$ is the flow matching time, $\bs \in \R^{k \times d}$ are $k$ \OursABV representing the shape $\cS$, and $\theta$ is the parameters of a neural network $v$.
The \OursABV $\bs$ are outputs of a tokenizer $\mu_\theta(\cS)$, which is jointly learned with the flow matching model to embed all necessary information about $\cS$ into $\bs$ to fit $p_{\cS}(x)$.
To input information $\cS$ to $\mu$, we sample a point cloud containing $m$ points on $\cS$ --- this enables us to train both $\mu$ and $v$ with only point clouds.
Specifically, given a dataset containing $N$ point clouds, $\cX^1, \dots, \cX^N$, where $\cX^i$ contains $n$ \iid samples $x_1^i, \dots x_n^i \sim p_{\cS^i}(x)$ of shape $\cS^i$ and $n \, {\gg} \, m$, we maximize the variational lower bound of the log-likelihood of the empirical distribution:
\begin{align}
 &	\max_{\theta} \ \bbE_{\cS}  \ \bbE_{x \sim p_{\cS}(x )} \  \log p_\theta(x | \cS)  \\
 &	\approx   \bbE_{\cS}  \ \bbE_{x \sim p_{\cS}(x )} \  \log  \int_\bs p_\theta(x , \bs | \cZ) \dd{\bs}  \label{eq: approx shape with point cloud}
 \\
 &	=    \bbE_{\cS}  \ \bbE_{x \sim p_{\cS}(x )} \  \log  \int_\bs  p_\theta(x | \bs) p_\theta(\bs | \cZ) \frac{q_\theta(\bs | \cY)}{q_\theta(\bs | \cY)}\dd{\bs}   \\
 &	\ge  \bbE_{\cS}  \ \bbE_{x \sim p_{\cS}(x )} \  \bbE_{\bs \sim q(\bs | \cY)} \log  p_\theta(x | \bs)  {-} KL(q_\theta(\bs | \cY) || p_\theta(\bs | \cZ)), \label{eq: diff}
\end{align}
where $\cY$ and $\cZ$ are independently sampled point clouds containing $m$ points.
The approximation in \Cref{eq: approx shape with point cloud} is from using $\cZ$ as $\cS$ and is controlled by the density of $\cZ$ (\ie, number of points, $m$).
We apply Jensen's inequality at \Cref{eq: diff}.  %
Since all models are jointly trained, we use $\theta$ to represent all learnable parameters.
We use flow matching (\ref{eq: flow matching loss}) to learn $p_\theta(x | \bs)$, we parameterize $q_\theta(s | \cY)$ as a Gaussian distribution $\cN(\bs; \mu_\theta(\cY), \sigma^2 I)$ and $p_\theta(\bs | \cZ)$ as $\cN(\bs; \mu_\theta(\cZ), \sigma^2 I)$. 
Under this parameterization, the KL divergence in \Cref{eq: diff} is reduced to $\frac{1}{\sigma^2} \| \mu_\theta(\cY) - \mu_\theta(\cZ) \|^2$.  
This is intuitive as two point clouds sampled from the same shape should produce similar \Ours.
To regularize the shape-token space, we also add a KL-divergence $KL(q_\theta(\bs | \cY) \; || \;  p(\bs))$, where $p(\bs)$ is the prior distribution of $\bs$, an isometric Gaussian distribution.
We utilize a weighted sum of the KL divergence terms ($10^{-3}$ for \Cref{eq: diff} and $10^{-4}$ for the prior term) in the training objective and set $\sigma=10^{-3}$ empirically.

\paragraph{Architecture.}
The architecture of the shape tokenizer $\mu(\cdot) \,{\rightarrow}\, \bs$ and the flow-matching velocity estimator $f$ are illustrated in \Cref{fig: architecture} and detailed in \Cref{supp sec: arch}.
The shape tokenizer has a similar architecture as PerceiverIO~\cite{perceiverio}.
We learn an array of initial queries that retrieve information from the input point cloud representing the shape with a cross attention block.
Two self-attention blocks are added after each cross attention block. %
A linear layer projects the output of the last self attention block to \OursABV.

The velocity estimator $v$ takes the \OursABV $\bs$ and a 3D location $x$ as input and outputs the estimated 3D velocity at $x$.
We use $x$ as query of the first cross attention (and $\bs$ as key and value).
The flow matching time $t$ is supplied through adaptive layer normalization.
Note that the velocity decoder is intentionally simple---no self attention is in the velocity decoder, making the velocity estimation at each 3D location independent to each other, modeling $p(x | \bs)$ as a 3D distribution (instead of a joint distribution of $n$ points).

\subsection{Properties}
\label{sec: analysis}

We analyze modeling 3D shapes with 3D flow matching.

\paragraph{Zero-shot surface normal estimation.}
Since $p(x|\cS)$ is a 3D delta function lying on the surfaces, its gradient direction \wrt $x$, aligns with the surface normal when $x$ is on the surfaces.
This allows us to estimate surface normal without access to any training data containing ground-truth surface normal.
We use the formula derived by \citet{sit} to convert velocity to score function and result in
\begin{equation}
	\hat{n}(x) = \frac{\grad_x{\log p_1(x | \bs)} }{ \| \grad_x{\log p_1(x | \bs)} \| } = \text{normalize}(\alpha_1 v(x; t \,{\rightarrow}\, 1) - \dot{\alpha}_1 x).
	\label{eq: normal}
\end{equation}
The expression motivates us to choose the generalized variance preserving path ($\alpha_t = \sin(\frac{\pi}{2}t)$ and $\sigma_t = \cos(\frac{\pi}{2}t)$)~\cite{sit, albergo2023building} such that  $\dot{\alpha}_1 = 0$ and the decoder directly estimates the normal direction when $t = 1$.

\begin{figure}[t]
	\centering
	\includegraphics[width=\linewidth]{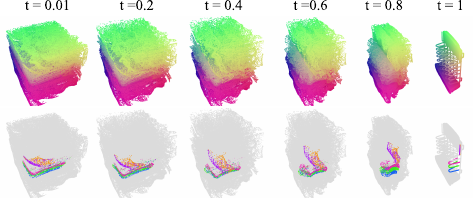}
	\vspace{-4mm}
	\caption{The ODE integration trajectory maps xyz (data) to uvw (noise). Mesh credits~\cite{downs2022google}.}
	\label{fig: gso uvw 00001}
\end{figure}

\paragraph{UVW mapping.}
Flow matching has non-intersecting ODE integration trajectories and bijective mapping between the initial noise space and the 3D shape space as our velocity estimator is uniformly Lipschitz continuous in $x$ and continuous in $t$~\cite{chen2018neural}.
This means that given any 3D location in the 3D space (\ie, xyz), we can traverse the ODE trajectory (\ref{eq:prob_flow}) back to a unique location in initial noise 3D space (that we call uvw to differentiate from xyz).
Additionally, since the trajectories do not intersect, the mapping varies smoothly.
Inspired by the property, we choose to use a uniform distribution within $[-1, 1]$ as our initial noise distribution $\e(\epsilon)$.
This allows us to map each xyz to a location in a uvw-cube.
One example is shown in \Cref{fig: gso uvw 00001}.
We also use the property to color the sampled point clouds by their initial uvw locations (rgb = (uvw + 1) / 2).  As can be seen in the figures, the uvw mapping varies smoothly across xyz.
We think the uvw-mapping is an interesting property and worth noting in the paper, as it is automatically discovered by shape tokenization and resembles the UV-mapping technique used for texture interpolation.
This capability is unique of our flow matching decoder and not possessed by occupancy/SDF decoders used in existing works~\cite{zhao2023michelangelo, zhang20233dshape2vecset}.

\paragraph{Surface likelihood.}
Flow matching enables using the instantaneous change of variable~\cite{chen2018neural} to calculate the exact log-likelihood $\log p(x | \bs)$ at any 3D location, enabling us to estimate the probability of $x \, {\in} \, \cS$ by integrating an ODE.
Since our distribution is 3-dimensional, we can calculate the exact divergence with automatic differentiation with little cost instead of using a trace estimator as by \citet{song2021scorebased}.
The capability to evaluate log-likelihood at any location is useful for removing noisy points, \eg, due to the finite number of steps used for the ODE integration.
We also notice that in practice we can get good estimation of the log-likelihood by integrating the ODE for log-likelihood estimation with much fewer steps (\eg, 25).
In the paper, we use the technique to filter point clouds sampled from $p(x|\bs)$ to filter the stray/noisy points caused by numerically integrating the ODE when sampling $p(x|\bs)$.
Please see pre-filtered point clouds in the supplemental material.

%% file: sec/4_exp.tex
\section{Experiments}

We evaluate shape tokenization from two perspectives.
In \Cref{sec: reconstruction}, we assess the geometry information preserved in \OursABV by comparing with ground-truth shapes.
From \Cref{sec: exp clip} to~\ref{sec: exp rendering}, we apply \OursABV as the 3D representation in three applications, 3D CLIP (\Cref{sec: exp clip}), 3D shape generation (single-image-to-3D or unconditional generation) (\Cref{sec: exp lfm}), and estimation of ray-shape intersection (\Cref{sec: exp rendering}).
Our goal is to demonstrate a variety of capabilities enabled by \OursABV and further motivate future work.

\input{sec/4_0_reconstruction_v2}

\input{sec/4_2_clip}

\input{sec/4_1_lfm}

\begin{figure}[t]
	\centering
	\includegraphics[width=\linewidth]{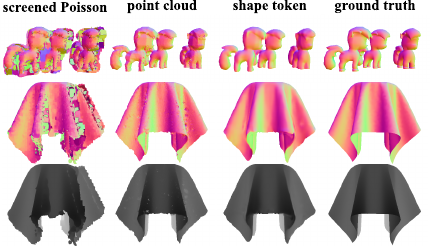}
	\vspace{-7mm}
	\caption{Given a point cloud containing 16,384 points (xyz only), camera pose and intrinsics, we estimate the normal (top two rows) and depth (bottom row) of the intersection point of each ray separately. (Left) applies screened Poisson reconstruction and performs intersection on the reconstructed mesh. (Middle left) directly feeds input point cloud to the pointersect model.  (Middle right) convert point cloud to \OursABV and feed to modified pointersect model. Mesh credits~\cite{shakiller_pony, caitlin_test_fabric}.}
	\label{fig: pointersect}
\end{figure}

\input{sec/4_3_graphics}

%% file: sec/4_0_reconstruction_v2.tex
\subsection{Reconstruction}
\label{sec: reconstruction}

We first train shape tokenizers on ShapeNet dataset~\cite{shapenet2015}, and later scale our training data to Objaverse dataset~\cite{objaverse}. 
We follow LION's evalution~\cite{vahdat2022lion} and measure chamfer distance with point clouds of scale [-0.5, 0.5].

\paragraph{ShapeNet.} ShapeNet dataset~\cite{shapenet2015} contains 55 classes of objects.  For fair comparison, we use the same training data as Pointflow~\cite{yang2019pointflow} and LION~\cite{vahdat2022lion}, including point cloud normalization and train-evaluation splits.
The training set contains 35,708 point clouds, and the test set contains 10,261 point clouds.
All point clouds contain 15,000 points.
We randomly sample 4,096 points without replacement from the 15,000 points, and we use the first 2048 points as input to the tokenizer and the rest as the reference when computing symmetric Chamfer Distance (CD).
We use 500 uniform steps of Heun's $2^{\text{nd}}$ order method~\cite{karras2022elucidating} to sample.  %

We compare with Pointflow~\cite{yang2019pointflow}, LION~\cite{vahdat2022lion}, and 3DShape2VecSet~\cite{zhang20233dshape2vecset}.
PointFlow represents a shape as a latent vector of 512 dimensions, LION represents a shape as a 128-dimensional global latent and a local latent of 8192 total dimensions, and 3DShape2VecSet has a total latent dimension of 4096.
We also conduct ablation study on our method by training various tokenizer with different total latent dimensions (number of tokens multiplied by their dimension).  
All methods take 2048 input points.  %

As shown in \Cref{fig: architecture} and \Cref{table: shapenet reconstruction} in appendix, while using higher latent dimensions generally leads to better reconstruction, shape tokenizers achieve better trade-off than existing methods. 
For example, it achieves similar chamfer distances while being 16$\times$ more compact than LION.
We think this is because we model a 3D distribution whereas they model a $2048{\times}3$-dimensional distribution, making our method data efficient.
Our ablation also shows improvements of chamfer distances as we increase the total latent dimension of \OursABV.

\begin{table}[tb]
	\caption{\footnotesize Reconstruction error on Objaverse and GSO datasets. Chamfer distances are computed on 8192 points and are of unit $10^{-4}$.  The first row block are models trained on ShapeNet, \OursABV in the second block are trained on Objaverse, and TRELLIS trained on ObjaverseXL and three other datasets. * indicates normal estimated by Open3D.}
	\label{table: reconstruction objaverse}
	\vspace{-3mm}
	\centering

    \begin{adjustbox}{max width=\linewidth}
		\begin{tabular}{llcccc}
			\toprule
			&
			\multirow{2}[4]{*}{\makecell{latent}} &
			\multicolumn{2}{c}{Objaverse}   &
			\multicolumn{2}{c}{Google scanned objects}
			\\
			&
			&
			CD $\downarrow$ &  %
			normal ($^\circ$)$\downarrow$&
			CD $\downarrow$ &
			Normal  ($^\circ$) $\downarrow$
			\\
			\midrule

			{Michelangelo} &
			{$512{\times}64$}&
			{${27.5} \pm 47.7$} &
			{${28.7} \pm 13.7$} &
			{${10.8} \pm 17.4$} &
			{${18.3} \pm 11.3$}
			\\

			{3DShape2VecSet} &
			{$512{\times}8$}&
			{${6.5} \pm 13.6$} &
			{$\mathbf{20.1} \pm 12.0$} &
			{${4.1} \pm 11.0$} &
			{$\mathbf{12.2} \pm 8.0$}
			\\
            
			{ST-shapenet} &
			{$256{\times}16$}&
			{${3.7} \pm 2.1$} &
			{$38.3 \pm 10.3$} &
			{${3.2} \pm 1.6$} &
			{$29.0 \pm 12.6$}
			\\
            
			{ST-shapenet} &
			{$512{\times}16$}&
			{$\mathbf{2.8} \pm 1.7$} &
			{${32.1} \pm 11.2$} &
			{$\mathbf{2.6} \pm 1.2$} &
			{${23.2} \pm 12.0$}
			\\
            
			\midrule
			{TRELLIS} &
			{${\sim}20000{\times}11$}&
			{$2.0 \pm 1.3$} &
			{$\mathbf{15.4} \pm 10.9$} &
			{$2.2 \pm 0.90$} &
			{$\mathbf{10.0} \pm 4.92$}
			\\
            
			{ST-objaverse} &
			{$1024{\times}8$}&
			$1.8 \pm 1.2$ &
			$22.5 \pm 10.2$ &
			$2.0 \pm 0.85$ &
			$15.1 \pm 8.26$
			\\
            
			{ST-objaverse} &
			{$1024{\times}16$}&
			$\mathbf{1.6} \pm 1.1$ &
			$19.0 \pm 8.84$ &
			$\mathbf{1.9} \pm 0.83$ &
			$13.9 \pm 6.84$
			\\

			\midrule
			\multicolumn{2}{l}{real 2048 points}  &
			$3.2 \pm 2.5$ &
			$25.3 \pm 11.1$* &
			$4.0 \pm 1.8$ &
			$17.6 \pm 8.74$*
			\\
			\multicolumn{2}{l}{real 8192 points}  &
			$1.3 \pm 1.1$ &
			$20.5 \pm 9.95$* &
			$1.6 \pm 0.77$ &
			$14.3 \pm 7.68$*
			\\
			\bottomrule
		\end{tabular}
	\end{adjustbox}

	\vspace{-0.3cm}

\end{table}

\begin{table}[t]
    \setlength{\tabcolsep}{0.1cm}
	\caption{\footnotesize Reconstruction error on room scenes. Chamfer distances are computed with 8192 points and are of unit $10^{-4}$.}
	\label{table: reconstruction room}
	\vspace{-3mm}
	\centering
	\begin{adjustbox}{max width=\linewidth}
		\begin{tabular}{ccccccc}
			\toprule
			\multicolumn{3}{c}{ARKitScenes~\citep{baruch1arkitscenes}}   {\hspace{8mm}} &
			\multicolumn{4}{c}{HM3D~\citep{ramakrishnan2habitat}} 
			\\
            \cmidrule(lr){1-3} \cmidrule(lr){4-7}
			\makecell{ST-shapenet \\ $512 {\times} 16$} &
			\makecell{ST-objaverse \\ $1024 {\times} 16$} &
			\makecell{real\\8192} &
			\makecell{\cite{zhang20233dshape2vecset} \\ $512 {\times} 8$} &
			\makecell{ST-shapenet \\ $512 {\times} 16$} &
			\makecell{ST-objaverse \\ $1024 {\times} 16$} &
            \makecell{real\\8192} 
			\\
			\midrule
			$2.6 \pm 0.94$  & %
			$\mathbf{0.92} \pm 0.42$  & %
			$0.74 \pm 0.38$  & %

			$12.2 \pm 1.8 $  &  %
			$5.1 \pm 2.0$  & %
			$\mathbf{2.8} \pm 1.3$  &%
			$2.3 \pm 1.1$   %
			\\
			\bottomrule
		\end{tabular}
	\end{adjustbox}
	\vspace{-2mm}

\end{table}

\paragraph{Objaverse.}  We train shape tokenizers on the Objaverse dataset~\cite{objaverse}, which contains 800k meshes with a wide variety of 3D shapes.
Unlike existing methods that need watertight meshes for training (\eg, to compute signed distance functions or occupancy \cite{zhang2024clay, zhao2023michelangelo, wu2024direct3d}), our method only requires point clouds.
This significantly simplifies our training and enables us to utilize all meshes in the training split and do not perform any preprocessing (\eg, smoothing the meshes, making the meshes watertight, \etc) besides box-normalization to [-1, 1].
We \iid sample 200k points uniformly on mesh surfaces to create a dataset of point clouds.
We randomly select 640k meshes for training.

We evaluate the shape tokenizers on 400 held-out meshes in Objaverse and the entire Google Scanned Objects (GSO) dataset~\cite{downs2022google}, which contains 1032 meshes.
We sample without replacement $16384 + n$ points from each point cloud, the 16384 points are used as input to the tokenizer and the $n$ points are used as reference when computing chamfer distance.
We also evaluate estimated surface normal by computing the angle between the estimated and the ground-truth normal. 
Since an estimated 3D point (which the estimated normal attached to) may not lie on the actual surface, the ground-truth normal is that of the closest point in a densely sampled real point cloud (containing 200k points).

We compare with models trained on ShapeNet (including \OursABV-shapenet, Michelangelo, and 3DShape2VecSet) and \OursABV trained on Objaverse (\OursABV-objaverse), to demonstrate the effect of dataset scaling.\footnote{When evaluating Michelangelo and 3DShape2VecSet on Objaverse, we find that $\sim$1\% of the results have large errors, whereas they do not have the issue on GSO. We hypothesize that this is due to the models are trained with watertight meshes which Objaverse meshes may violate. Thus, for the two methods we remove their worst 2\% of results on Objaverse. For \OursABV we report with all results (\ie, no filtering).}  
We also evaluate a concurrent work, TRELLIS \cite{xiang2024structured}.
Since we focus on geometry, we compare with TRELLIS's pretrained mesh decoder.
Note that the comparison with TRELLIS is just for completeness---TRELLIS utilizes additional input information (150 multi-view images and DINOv2-large feature) and supervision (normal maps) than ours (point clouds only).
We also create an oracle baseline by independently sampling real point clouds from the surfaces (shown as real \# points in \Cref{table: reconstruction objaverse}).
When the number of points is the same as the reference point cloud, it provides the lower-bound on chamfer distance, and when the number points is smaller than the reference point clouds, we randomly sample with replacement to match the number of points in the reference point cloud.  
For normal estimation, we provide another baseline that estimates normal by fitting planes locally using Open3D~\cite{Zhou2018} on the real point cloud.  
This is a standard method when point clouds do not contain normal.

The results are shown in \Cref{table: reconstruction objaverse}, from which a few observations can be made. 
First, scaling training data improves \OursABV results---\OursABV trained on Objaverse achieves better performance than those trained on ShapeNet, despite having the same total latent dimension ($1024 \times 8$ vs $512 \times 16$). 
Second, shape tokenizers trained on Objaverse generalize well to unseen shapes in GSO, achieving chamfer distances close to the upper bounds.
Third, the zero-shot normal estimation of \OursABV-shapenet performs poorly on Objaverse. We hypothesize this is due to the flow matching model cannot generalize from the small size and smooth surfaces of ShapeNet to Objaverse.  In comparison, by leveraging prior knowledge of 3D (learning to reconstruct occupancy field and extract meshes with marching cube), 3DShape2VecSet achieves good normal estimation. 
However, by scaling training data of \OursABV, we can significantly improves the performance and bridge the gap on zero-shot normal estimation.
Last, our method achieves similar chamfer distances as TRELLIS, which is one of the state-of-the-art methods, despite using less information.  
Our zero-shot normal estimation is also close to the performance of TRELLIS, which is supervised by normal maps during training. 
\Cref{fig: gso reconstruction} in appendix shows examples of our sampled point clouds on GSO. Please see the supplemental material for more examples. 
The full ablation study on the latent dimension of \OursABV-shapenet and \OursABV-objaverse is in \Cref{table: reconstruction objaverse full} in the appendix.

\paragraph{Room scenes.}  We evaluate shape tokenizers on 100 room scenes from ARKitScenes (lidar-scanned point clouds)~\citep{baruch1arkitscenes} and 13 scenes from HM3D~\citep{ramakrishnan2habitat}.
The metrics is reported in \Cref{table: reconstruction room}, and \Cref{fig: hm3d} in appendix shows the reconstruction results.
As expected, \OursABV-objaverse generalizes better to room scenes than \OursABV-shapenet.
Despite trained on object-centric data, details like room structures, sofas, and tables are preserved; however, we do notice loss of details in the reconstructed rooms.
We hypothesize that the cause is the limited number of input points (16,384) to represent complex scenes like houses with several floors and rooms.
Extending to complex scenes is an interesting future work.

\begin{figure*}[t]
	\centering
	\includegraphics[width=\linewidth]{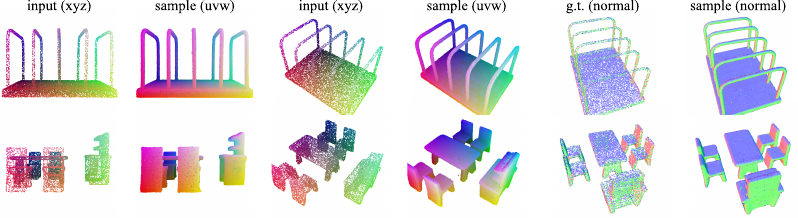}
	\vspace{-7mm}
	\caption{Reconstruction, densification, and normal estimation of unseen point clouds in GSO dataset. For each row, we are given a point cloud containing 16,384 points (xyz only), we compute \OursABV and \iid sample the resulted $p(x|\bs)$ for 262,144 points. Different columns render the input and the sampled point clouds from different view points. Indicated by the label in the parenthesis, we color the input points according to their xyz coordinates and the sampled points according to their initial noise's uvw coordinates and their estimated normal (last two columns).  Note that we do not provide normal as input to the shape tokenizer. Mesh credits~\cite{downs2022google}.
	}
	\label{fig: gso reconstruction}
\end{figure*}

%% file: sec/4_2_clip.tex
\subsection{Integrating \OursABV to CLIP alignment}
\label{sec: exp clip}

3D-CLIP aims to align 3D shape embeddings with image and text embeddings of a pretrained CLIP model~\cite{ilharco_gabriel_2021_5143773}.
The shape encoder takes a point cloud as input and outputs an embedding.
We replace the shape encoder (PointBERT) of an existing 3D-CLIP pipeline, OpenShape \cite{liu2024openshape}, with our shape tokenizer ($1024 \times 16$) and an MLP.
The MLP takes the concatenated \OursABV as an input vector and has 4 layers of feature dimension 4096, and finally a linear layer output the embedding of dimension 1280.
Note that we only train the MLP, and thus we are able to use a large batch size (600 per GPU).
For apple-to-apple comparison, we use the same training recipe and datasets as OpenShape,~\ie,~a combined dataset of Objaverse~\citep{objaverse}, ShapeNet~\citep{shapenet}, 3D-FUTURE~\citep{fu20213d}, and ABO~\citep{collins2022abo}.
We also use the same text captions as OpenShape.
The models are trained for 2 weeks using 8 A100 GPUs.
We evaluate the learned CLIP-aligned shape embedding with zero-shot text classification.
We compare with the pretrained OpenShape+PointBERT that takes xyz as input (as ours).
As can be seen in \Cref{table: clip}, the model trained with \OursABV as the 3D representation achieves better performance as OpenShape that uses a specifically trained PointBERT encoder.

\input{tables/combined_img_cond_clip}

%% file: tables/combined_img_cond_clip.tex
\begin{table}[t]
    \begin{minipage}{0.43\linewidth}
        \setlength{\tabcolsep}{0.25cm}
    	\caption{\footnotesize Zero-shot text classification on Objaverse-LVIS.
        }
    	\label{table: clip}
    	\vspace{-3mm}
    	\centering
    	\begin{adjustbox}{max width=\linewidth}
    		\begin{tabular}{lcc}
    			\toprule
                \makecell[c]{shape encoder}
                &
    			top-1   &
    			top-5
    			\\
    			\midrule
    			PointBERT &  %
    			42.6 &
    			73.1
    			\\
    			ST  &
    			\textbf{48.4} &
    			\textbf{75.5}
    			\\
    			\bottomrule
    		\end{tabular}
    	\end{adjustbox}
    \end{minipage}
    \hspace{3mm}
    \begin{minipage}{0.55\linewidth}
        \setlength{\tabcolsep}{0.1cm}
    	\caption{\footnotesize Single-image-conditioned generation on Objaverse.}
    	\label{table:objaverse}
    	\vspace{-3mm}
    	\centering
    	\begin{adjustbox}{max width=\linewidth}
    		\begin{tabular}{lccc}
    			\toprule
    			Model & ULIP-I $\uparrow$ & P-FID $\downarrow$ & P-IS $\uparrow$ \\
    			\midrule
    			Shap-E \cite{jun2023shap} & 0.13 & - & - \\
    			Michelangelo \cite{zhao2023michelangelo} & 0.19 & - & - \\
    			CLAY \cite{zhang2024clay} & 0.21 & 0.99 & - \\
    			ST (ours) & \textbf{0.32} & \textbf{0.77} & 11.4 \\
    			\bottomrule
    		\end{tabular}
    	\end{adjustbox}
    	\vspace{-0.4cm}
    \end{minipage}
\end{table}

%% file: sec/4_1_lfm.tex
\subsection{Integrating \OursABV to 3D generation}
\label{sec: exp lfm}

\begin{figure*}[t]
	\centering
	\includegraphics[width=\linewidth]{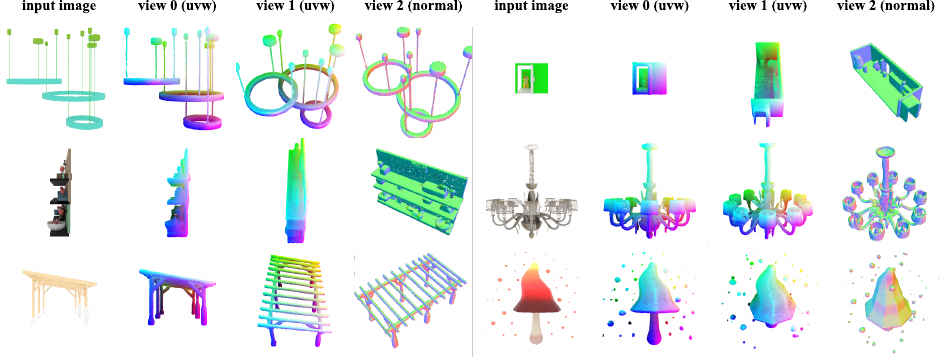}
	\vspace{-6mm}
	\caption{Single-image to 3D point cloud results on unseen meshes in Objaverse. We color the points with RGB color that indicates the original location of the point in the initial noise space.
			Mesh credits~\cite{fedomo_ru_svetilnik_3885_25la,sketchingsushi_cali_garden,binkley_spacetrucker_galactic_truckstop_restrooms,fedomo_ru_lyustra_2054_10p,anyaachan_red_glowing_mushroom,martinice_group_op220667}.
	}
	\label{figure: 3d}
    \vspace{-0.3cm}
\end{figure*}

To demonstrate \OursABV is compatible with generative models, we train Latent Flow Matching (LFM) models that generate \OursABV.
First, we train an unconditional LFM on ShapeNet dataset (55 classes)~\cite{shapenet2015} and an image-conditioned LFM on Objaverse dataset~\cite{objaverse}.
The flow matching model architecture is based on the Diffusion Transformer (DiT)~\cite{dit}. 
\Cref{sec: details of lfm} provides details about model architectures, training recipe, and flow matching sampling.

We compare with LION \cite{vahdat2022lion}, 3DShape2VecSet \cite{zhang20233dshape2vecset}, and DPF \cite{zhuang2023diffusion}, and evaluate the results with metrics used by LION. 
Specifically, LION is a latent diffusion model that models the joint distribution of a fixed size point set.
DPF is our implementation of \citep{zhuang2023diffusion}. 
The model shares similar architecture and number of parameters as ours, but it is trained end-to-end to directly models the coordinates of 3D points with flow matching.  It is a strong baseline for point-cloud generation.
3DShape2VecSet only provides a class-conditioned pretrained model; to compare with other unconditional models, we provide the model with class labels whose  distribution matching that in the training data.
Our ShapeNet LFM model takes 32 \OursABV of dimension $64$; it has $\sim$110M parameters, which is similar to the sizes of LION and DPF.
We measure Minimum Matching Distance (MMD), Coverage (COV), and 1-Nearest Neighbor Accuracy (1-NNA)~\cite{yang2019pointflow}, using the same reference point clouds used by \citet{vahdat2022lion}.
For each method, we sample 1000 point clouds containing 2048 points.
As can be seen in \Cref{table:shapenet}, our model achieves better performance than LION, potentially contributed by our more compact latent space.
Our model also achieves competitive performance with DPF, that learns end-to-end the point distribution without a separately learned encoder.

\begin{table}[t]
	\caption{\small Unconditional generation on ShapeNet. For MMD-CD, the unit is $10^{-3}$, and that for MMD-EMD is $10^{-2}$. *3DShape2VecSet is class-conditional, and we sample class distribution matching that in training data.}
	\label{table:shapenet}
	\vspace{-3mm}
	\centering
	\begin{adjustbox}{max width=\linewidth}
		\begin{tabular}{lcccccc}
			\toprule
            & \multicolumn{2}{c}{MMD $\downarrow$}&\multicolumn{2}{c}{COV $\uparrow$ (\%)}&\multicolumn{2}{c}{1-NNA $\downarrow$ (\%)} \\
            \midrule
			Model & CD & EMD & CD & EMD & CD & EMD \\
			\midrule
			3DShape2VecSet* \cite{zhang20233dshape2vecset} & 3.51  &          2.18    &      \textbf{51.0}   &        52.0    &      59.2  &         60.6 \\
			LION \cite{vahdat2022lion} & 3.43 & 2.10 & 48.0 & 52.2 & 58.3 & 57.8 \\
            DPF \cite{zhuang2023diffusion} & 3.26 & 2.13 & 49.0 & 50.4 & \textbf{54.7} & 55.7 \\
            ST (ours)&  $\mathbf{3.25} {\pm 0.05}$ & $\mathbf{2.09} {\pm 0.01}$  & $50.4 {\pm 0.4} $  &  $\mathbf{53.0} {\pm 1.1 }$  &  $57.6 {\pm 1.2 }$ & $\mathbf{54.9} {\pm 0.9} $ \\
			\bottomrule
		\end{tabular}
	\end{adjustbox}
	\vspace{-0.3cm}
\end{table}

Next, we train an image-conditioned LFM that takes an image as input (represented by its DINOv2-small feature map) and generates \OursABV.  
The architecture and training details are in \Cref{sec: details of lfm}. 
\Cref{table:objaverse} shows the quantitative results of single-image-conditioned generation on Objaverse.
Due to the wide diversity of objects in Objaverse, we measure the quality of generated point clouds with ULIP-I~\cite{xue2024ulip}, P-FID~\cite{nichol2022point}, and P-IS~\cite{nichol2022point}.
We use the PointNet++ provided by \citet{nichol2022point} to measure P-FID and P-IS, which measures qualities of point clouds.
We use ULIP-2 \cite{xue2024ulip} to extract point-cloud embedding and measure cosine similarity with the conditioned image's CLIP embedding.
This evaluates the similarity between the generated point cloud and the input image.
Our LFM model generates \OursABV of dimension $1024 \times 16$ and has 612M trainable parameters.
As can be seen from the table, our model performs competitively to existing baselines.

\Cref{figure: 3d} shows examples of our single-image to point cloud results. Our model generates point clouds that highly resemble the conditioning images and have plausible 3D shapes.
Moreover, as shown in \Cref{figure: 3d multigen} in appendix, our model generates diverse samples when the conditioning is ambiguous (\eg, surfaces invisible in the conditioning images).
For example, given an image showing the outside of the room, the model generates rooms furnished differently inside. 
See \Cref{supp sec: more vis} for more generated results and videos.

\begin{figure}[t]
	\centering
	\includegraphics[width=\linewidth]{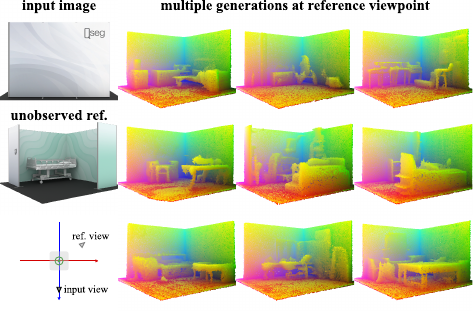}
	\vspace{-6mm}
	\caption{We generated 9 point clouds independently from the same input image (top left image). We provide the rendered image of the meshes at the same viewpoint as a reference (middle left). No human selection was conducted.  Note that the model does not observe the reference images. Mesh credits~\cite{exhibitbook_qseg_isolation}. 
    }
	\label{figure: 3d multigen}
\end{figure}

%% file: sec/4_3_graphics.tex
\subsection{Integrating \OursABV to neural rendering}
\label{sec: exp rendering}

Ray-shape intersection is an important operator in graphics.  It is also a difficult task when the 3D shape is given as a point cloud.
We demonstrate the capability of estimating the intersecting location with a neural network by representing the input point cloud with \OursABV.
Specifically, we train a transformer consisting of 4 cross-attention blocks.
The first cross attention takes the Plucker embedding of a single ray as query and attends to the \OursABV.
In other words, individual rays are independently processed.
A final linear layer outputs estimations of whether the ray hits any surface, the ray-traveling distance to the first intersection point, and its surface normal.
Given a camera pose and intrinsics, we process rays corresponding to each pixel individually and rasterize depth and normal map images.
We compare with an existing method, Pointersect~\cite{chang2023pointersect}, that directly uses the input point cloud and a ray as input of a transformer to estimate intersection points.
We train our model with the same loss function as Pointersect.
The results are shown in \Cref{table: pointersect} and \Cref{fig: pointersect}.
With \OursABV as its 3D representation, the model is able to estimate a smoother normal map, robust to local variations of point clouds.

\begin{table}[t]
	\caption{Ray-shape intersection on Objaverse.}
	\label{table: pointersect}
	\vspace{-3mm}
	\centering
	\begin{adjustbox}{max width=\linewidth}
		\begin{tabular}{lccc}
			\toprule
			&
			Poisson  &  %
			Pointersect &  %
			ST (ours) + Pointersect
			\\
			\midrule
			Depth (RMSE) $\downarrow$ &
			$0.16 \pm 0.15$  &  %
			$0.064 \pm 0.066$ &  %
			$\mathbf{0.053}\pm 0.055$   %
			\\
			Normal (angle ($^\circ$)) $\downarrow$ &
			$26.0 \pm 14.1$  &  %
			$17.7 \pm 12.3$ &  %
			$\mathbf{15.7} \pm 12.3$   %
			\\
			Hit (acc ($\%$)) $\uparrow$ &
			$91.1 \pm 10.6$  &  %
			$\mathbf{99.4} \pm 1.20$ &  %
			$99.3 \pm 1.19$   %
			\\
			\bottomrule
		\end{tabular}
	\end{adjustbox}
	\vspace{-0.3cm}
\end{table}

%% file: sec/5_discussion.tex
\section{Discussion}
\label{sec: discussion}

Shape tokenization is a novel 3D shape representation that is purely data-driven and designed to be consumed by ML models.
It lies on the opposite end of the spectrum from most existing 3D representations, which explicitly model geometry (\eg, meshes, SDFs) or rendering formulations (\eg, 3D Gaussians, NeRF).
Despite being data-driven, we show that \Ours possess properties that are tightly connected to 3D geometry, such as surface normals and UVW mapping.
The connection between flow matching and 3D geometry provides an interesting and new perspective of 3D representations.

\paragraph{Limitations.}
Current \Ours consider geometry only; extending to color is for future work.
We need to integrate ODEs when sampling point clouds, shape tokens, or computing log-likelihood, this means it takes longer times to generate a point cloud than feed-forward methods (see runtime analysis in \Cref{supp sec: runtime}).
Utilizing methods like distillation or advancement in diffusion models to improve sampling efficiency is also future work.
\OursABV is designed for ML models, not graphics, which however, often takes meshes as input. Extracting surfaces from \OursABV is left as future work (see more discussion in \Cref{sec: surface extraction}).

%% file: sec/99_appendix.tex
\input{tables/related_works_archive}

\xz{
\paragraph{The supplemental pdf} is organized as:
\begin{enumerate}
    \item \Cref{supp sec: more discussion} expands discussions on related works;
    \item \Cref{supp sec: impl} details implementations;
    \item \Cref{supp sec: more quant} provides more quantitative results;
    \item \Cref{supp sec: more vis} presents more visualizations.
\end{enumerate}
}

\section{More related works and discussions}\label{supp sec: more discussion}

\subsection{3D latent representation comparisons}\label{supp sec: extend related}

In the section, we expand our discussions in~\Cref{sec: related} on recent progress of 3D latent representations. 
We compile \Cref{table: related works}, which overviews recent latent 3D representations.
In the table, we categorize latent 3D representations by the entity they model, \eg, $p(xyz)$ like ours, $p(x_1, \dots, x_n)$ like LION~\cite{lion}, occupancy field like 3DShape2VecSet~\cite{zhang20233dshape2vecset}, \etc. 
We highlight the input to the encoder, the training datasets, total latent dimension (\ie, compactness),  whether the latent representation is fully continuous, and whether training the encoder requires extensive preprocessing.  

One observation from \Cref{table: related works} is that almost all recent methods use point clouds as input to the encoder.  
This may be due to the popularity of the transformer architecture and the continue natural of point clouds, making them a popular choice of input.

As stated in~\Cref{sec: method}, we model the probability density functions $p(xyz)$, following the perspective from DDPM-PointCloud~\cite{ddpm_pointcloud} and PointFlow~\cite{yang2019pointflow}, which use diffusion and continuous normalizing flow, respectively, to model the probability density functions. 
We rely on flow matching, which is closely related to diffusion and continuous normalizing flow but is simplified and has shown promising results on large scale datasets like videos~\cite{sepehri2024mediconfusion}. 
We also analyze the connection between 3D flow matching and geometry properties like surface normal and uvw mapping. 
One thing that is worth emphasizing is that like ours,  these works also only requires minimal preprocessing of 3D assets since we only need a set of 3D points.
The capability to train only with point clouds is an advantage, as point clouds can represent any 3D shapes (\eg, non-watertight, partial, or even broken ones from scans). 
Point clouds are also the output of most depth sensors and can thus be easily captured.

Point-E~\cite{nichol2022point} and LION~\cite{lion} also only require point clouds for training. 
However, they model the joint distribution of $k$ points, making the probability density a $3k$-dimensional function, as opposed to our $3$-dimensional function. 
We demonstrate similar reconstruction quality with $\frac{1}{16}$ total latent dimensions and better unconditional generation results than LION.

Many works train encoder-decoder to estimate occupancy or Signed Distance Function (SDF) values~\cite{zhang20233dshape2vecset, zhang20223dilg, zhao2023michelangelo, zhang2024clay, chen2024dora, yang2025pandora3d, wu2024direct3d, ren2024xcube, chou2023diffusion, yariv2024mosaic, li2025triposg, zhao2025hunyuan3d, hui2024make, siddiqui2024meshgpt, chen2024meshxl}. 
While modeling occupancy and SDF fields introduces inductive bias of 3D, it also inherently limits the type of 3D shapes and data these methods are able to train on.
Occupancy and SDF are only defined on watertight meshes. 
However, many objects in real world are not watertight, \eg, a thin piece of paper, tree leaves, a paper cup, \etc. 
Moreover, since most meshes made by artists are not watertight, preprocessing like closing and smoothening the meshes are needed when utilizing meshes in Objaverse or Objaverse-XL, as observed in multiple works~\cite{zhao2023michelangelo, zhang20233dshape2vecset, zhang2024clay, yang2025pandora3d}.
This hinders scaling up the training data. 

Some works utilize latent representations that are not fully continuous. 
For example, the sparsity patterns computed from input meshes~\cite{hui2024make, xiang2024structured}, quantization~\cite{zhang20223dilg, siddiqui2024meshgpt}, or hierarchical structure~\cite{ren2024xcube}. 
Depending on the applications, continuity of latent may or may not be a problem.  
For example, in generative applications, two- or multi-stage methods can be used~\cite{ren2024xcube, xiang2024structured}. 
However, if the application required taking input shape that is noisy (\eg, 3D shape denoising), the discontinuity in the input latent may cause difficulty in learning (for example, the sparsity pattern can miss the actual surface location due to noise).

Besides latent 3D representations, another line of work utilizes images and differentiable rendering to indirectly supervise explicit 3D representations utilized as part of the forward function~\cite{tang2025lgm,hong2024lrm,xu2024grm,xu2024dmvd,li2023instant3d,liu2023zero, gao2024cat3d,poole2023dreamfusion}. 
The focus of these works are the realism of the rendered images---3D representations and their distributions are not directly supervised or modeled, and thus are out of scope for this paper.

\subsection{Flow matching trajectory and UDF}
\label{sec: unsigned distance function}

Readers may compare or draw connections between the flow matching velocity field and unsigned distance functions (UDF). 
While the connection may seem plausible, we note the velocity / ODE trajectory should not be considered as an UDF. 
The flow-matching velocity moves probability density, which considers integrated surface area, whereas UDF only considers distance between two locations.  
We can provide an example where the velocity direction may differ from the UDF direction.
Suppose a scene contains with two spheres, one large at the left and one small at the right. 
An initial location is put in between the two spheres. 
Since the sphere on the left is larger and thus needs more probability density, depending on their size difference, flow matching may move us toward the left sphere to provide enough density (\ie, surface region) to the larger sphere, even when we are closer to the smaller sphere. 
However, in the case of UDF, it will only consider the distance to the closest surface.

\subsection{Surface extraction: future work}
\label{sec: surface extraction}

While we show that our sampled point clouds and estimated normal are close to the ground-truth, screened Poisson surface reconstruction~\cite{kazhdan2013screened} often fails to reconstruct surfaces from the sampled points. 

We hypothesize two potential reasons. 
First, screened Poisson surface reconstruction assumes  watertight shapes, whereas our model can generate non-watertight shapes. 

Second, our sampled point clouds are from a volumetric distribution, \ie, $p(xyz)$, where $xyz \in \R^3$.
As we model $p(xyz)$ with a neural network, which has a finite Lipschitz constant, the actual $p(xyz)$ that we sample from will not be a delta function on the 3D surface.
Instead, the density function will be highly concentrated on the surface but has a small width near the surface---imaging a 3D convolution between the delta function on the surface and a 3D gaussian kernel.
This means that when we sample the distribution, the sampled points will be close to but not exactly aligned to form a surface.
This volumetric nature makes utilizing Poisson surface reconstruction difficult, as they are not designed for volumetric point clouds and are sensitive to noise. 

We discuss two potential solutions for surface extraction.  
First, given a sampled point near surface, we can locally perturb the point to climb $p(xyz)$ to a local maximum by following the score function from the velocity decoder.
This is similar to non-max suppression and can be implemented with gradient ascent. 
Second approach is to directly learn a mesh decoder that takes shape tokens as input and output distance function or occupancy using end-to-end differentiable techniques like flexicube~\cite{shen2023flexicubes}. 
The similar technique has been used by TRELLIS to decode mesh from its sparse latent grid.

Surface extraction, mesh reconstruction and generation are vibrant research in computer graphics.  
As our goal is to study the use of flow matching in learning latent 3D representation to be consumed by machine learning models, not computer graphics, surface extraction from shape tokens is left as an interesting future work.

\section{Implementation details}\label{supp sec: impl}

\begin{enumerate}
    \item \Cref{sec: details of lfm} provides details about the latent flow matching models used to generate shape tokens;
    \item \Cref{supp sec: scaling} conducts ablation study on the model capacity of latent flow matching models of \OursABV;
	\item \Cref{supp sec: arch} explains architecture and training of Shape Tokenizer;
    \item \Cref{supp sec: arch neural rendering} details architecture and training of the neural rendering model;
    \item \Cref{supp sec: chamfer} details Chamfer distance formulation.
\end{enumerate}

\begin{figure}[t]
	\centering
	\includegraphics[width=\linewidth]{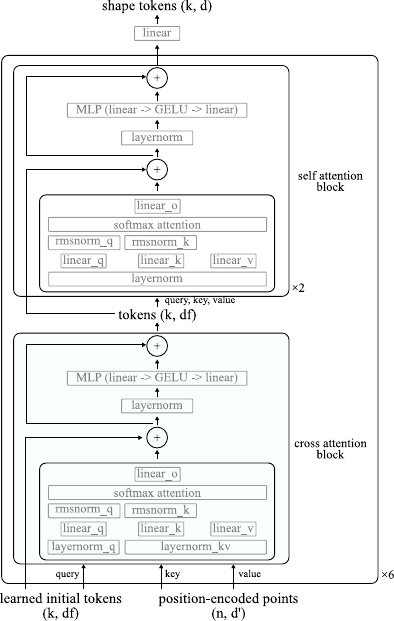}
	\vspace{-5mm}
	\caption{Architecture of shape tokenizer. Our main model for Objaverse uses  $n=16,384$, $k=1024$, $df = 512$, and $d=16$.  This results in $55.4$ million trainable parameters.
	For shape tokenizers trained on ShapeNet, we use $n=2048$, $k=32$, $df = 512$, and $d=64$, resulting in $54.9$ million trainable parameters.
	All multihead attention uses 8 heads.  The linear layers in MLP have expand and contract the feature dimension by 4 times.
	}
	\label{fig: arch tokenizer}
\end{figure}

\begin{figure}[t]
	\centering
	\includegraphics[width=\linewidth]{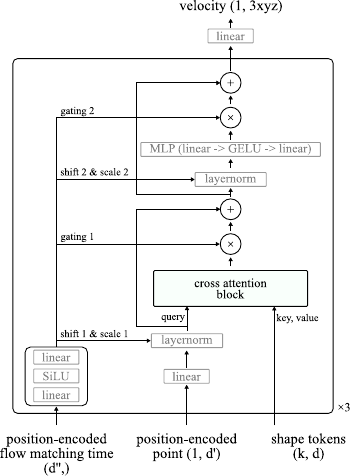}
	\vspace{-5mm}
	\caption{Architecture of flow-matching velocity estimator for shape tokenization. The model uses feature dimension 512,  and the number multihead attention is 8.  The linear layers in MLP have expand and contract the feature dimension by 4 times. The total number of trainable parameters is 8.72 millions and 8.87 millions for Objaverse and ShapeNet models, respectively.
	The neural rendering model uses the same architecture without the adaptive layer normalization with flow-matching time (\ie, it uses standard layer normalization layers).  It takes encoded ray as input and repeats the blocks 4 times.
	}
	\label{fig: arch flow}
\end{figure}

\subsection{Details of latent flow matching models}
\label{sec: details of lfm}

The section provides details about \Cref{sec: exp lfm}.

We train an unconditional Latent Flow-Matching (LFM) on ShapeNet dataset (55 classes)~\cite{shapenet2015} and an image-conditioned LFM on Objaverse dataset~\cite{objaverse}.
We use the same training splits as those used when training the shape tokenizers in \Cref{sec: method}.

\input{tables/3d_clip_full}

\begin{table*}[!t]
	\caption{Runtime (in seconds)}
	\label{table: runtime}
	\vspace{-3mm}
	\centering
	\begin{adjustbox}{max width=0.7\linewidth}
		\begin{tabular}{lcccc}
			\toprule
			&  \multicolumn{2}{c}{H100} & \multicolumn{2}{c}{A100}   \\
			& bfloat16 & float32 & bfloat16 & float32 \\
			\midrule
		    compute \Ours  &  0.022  & 0.028  &  0.025  &   0.047 \\
		    sample 16384 points from \Ours with 100 Euler steps  & 0.72   & 2.81 & 1.18  &  3.58 \\
		    sample shape tokens from single image with 100 Euler steps  & 6.76  & 23.12 & 9.20  & 33.56 \\
		    single image to point cloud   &  7.48  &  25.93 &   10.38  &  37.14 \\
			\bottomrule
		\end{tabular}
	\end{adjustbox}
	\vspace{-2mm}
\end{table*}

We build the latent flow matching models based on the Diffusion Transformer architecture (DiT)~\cite{dit} with AdaLN-single~\cite{chen2023pixart} and SwiGLU~\cite{shazeer2020glu}. 
The ShapeNet LFM model has 110 million trainable parameters. We train LFM model for Objaverse with various sizes, ranging from 31 million to 612 million trainable parameters (see \Cref{supp sec: scaling}). 
For the image-conditioned model, we use DINOv2-small~\cite{oquab2024dinov} to extract image features of each non-overlapping patch, and we encode the patch center rays' origins and directions with Fourier positional embedding and Plucker ray embedding~\cite{plucker2018analytisch}, respectively.
We also learn a linear layer to extract additional information from the image patches.
The DINO feature, the output of the linear layer, and the ray embedding of each patch are concatenated along feature dimension to form a vector $c$.
In each block, a cross-attention layer attends to $c$ of all patches to gather image information.

For each mesh in Objaverse, we render four images, each with 40 degrees field of view, $448 \times 448$ resolution, at 3.5 units on the opposite sides of x and z axes looking at the origin.
We train the models with AdamW optimizer \cite{loshchilov2017decoupled} with learning rate $10^{-4}$ with weight decay $0.01$.
The model is trained with batch size 128 for 200k iterations on ShapeNet and batch size 1024 for 1.2M iterations on Objaverse.
During sampling, we apply Heun's 2nd order method~\cite{karras2022elucidating} with uniform steps to sample both tokens (250 steps) and point clouds (100 steps).

\subsection{Scaling experiments on LFM models}
\label{supp sec: scaling}

\begin{figure*}[t]
	\centering
	\begin{subfigure}[t]{0.32\linewidth}
		\centering
		\includegraphics[width=\linewidth]{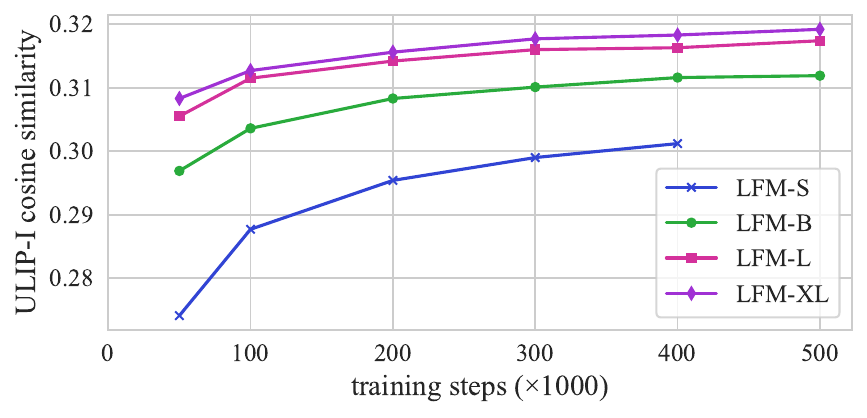}
	\end{subfigure}
	~
	\begin{subfigure}[t]{0.32\linewidth}
		\centering
		\includegraphics[width=\linewidth]{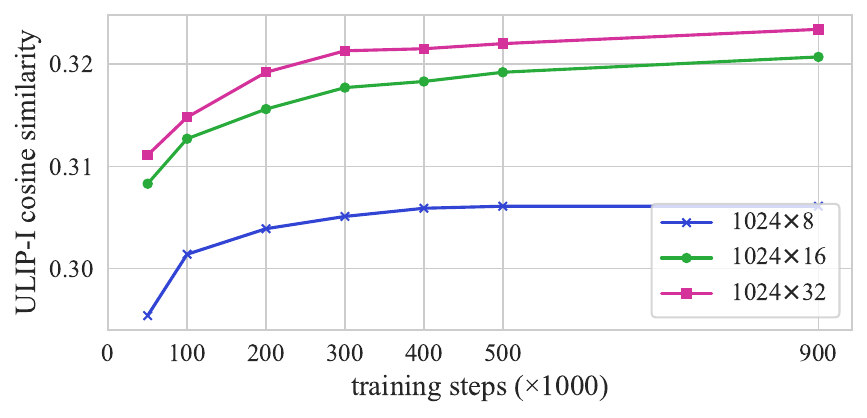}
	\end{subfigure}
	~
	\begin{subfigure}[t]{0.32\linewidth}
		\centering
		\includegraphics[width=\linewidth]{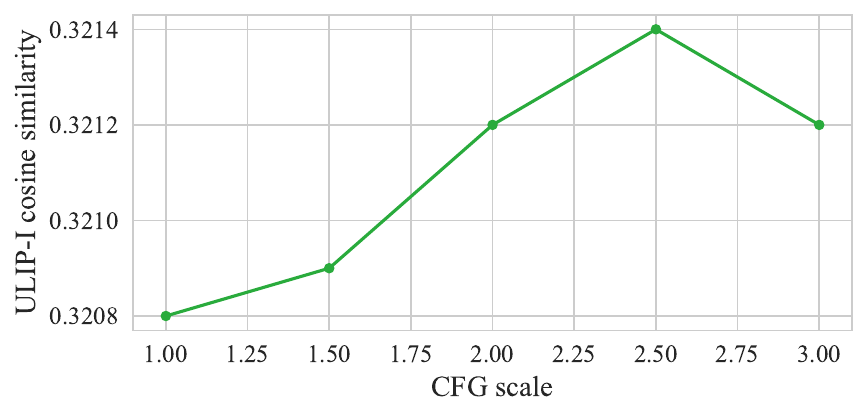}
	\end{subfigure}
	\vspace{-3mm}
	\caption{ULIP-I cosine similarities of (a) different model sizes, (b) different latent dimensions, and (c) different CFG scales.}
	\label{fig:clip}
\end{figure*}

In \Cref{fig:clip}, we demonstrate that the Latent Flow Matching model (LFM) trained on \Ours benefits from scaling, analogous to image tokenizers (e.g., SD-VAE \cite{ldm}).
We train LFM of various sizes: small (S), base (B), large (L), and extra-large (XL).
The number of trainable parameters for each model is (S) 31 million, (B) 121 million, (L) 423 million, and (XL) 612 million.  
As shown in \Cref{fig:clip}(a) and (b), the ULIP-I scores increase with the size of the models and dimension of the shape tokens.
Our model also supports classifer-free guidance (CFG).
\Cref{fig:clip}(c) illustrates how CFG scales affect the ULIP-I scores.

\subsection{Details of shape tokenizer}\label{supp sec: arch}

See \Cref{fig: arch tokenizer} for the detailed architecture of the shape tokenizer.  The main shape tokenizer trained on Objaverse has $55.4$ million trainable parameters, and the flow-matching velocity decoder has $8.7$ million trainable parameters.  
We use Fourier Positional embedding~\cite{vaswani2017attention} with $32$ logarithmic spaced frequencies from $2^0$ to $2^{12}$.
We change the dimension of the final linear layer to control the dimension of the \Ours.

See \Cref{fig: arch flow} for the detailed architecture of the velocity estimator paired with the shape tokenizer.
We use the same Fourier Positional embedding to encode the input xyz locations as that in the shape tokenizer.
We use Fourier positional embedding following by a MLP to encode flow-matching time.
The Fourier positional embedding uses $16$ logarithmic spaced frequencies from $2 \pi$  to $2^{16} \pi$, and the MLP has 2 linear layers (64 dimension) and SiLU activation function.

We train the shape tokenizer with AdamW~\cite{loshchilov2019decoupled} with $\beta_1 = 0.9$ and $\beta_2 = 0.98$.  No weight decay is used.  We use the learning rate schedule used by \citet{vaswani2017attention} with a warm-up period of 4000 iterations.
During the warm-up iterations, the learning rate increases to 2.8e-4, and it gradually decreases afterwards.
We train the shape tokenizers on 32 H100 GPUs for 200 hours (1.2M iterations).  We do not observe overfitting, since each point cloud contains a large number of \iid samples of $p_\cS(x)$.

\input{tables/recon_shapenet_full}

\input{tables/recon_objaverse_full}

\subsection{Details of neural rendering model}
\label{supp sec: arch neural rendering}

The neural rendering model uses the same architecture as the flow-matching velocity estimator above.
Without self-attention blocks, it processes individual rays independently.
We remove the adaptive layer normalization with flow-matching time (\ie, it uses standard layer normalization layers).
It takes encoded ray as input and repeats the blocks 4 times.
The ray is encoded as ray origin and direction. The coordinate of the ray origin is encoded with the same Fourier positional embedding as above.  The direction is encoded with Plucker ray representation~\cite{plucker2018analytisch}.  Additionally, we sample 32 points uniformly on the ray within the [-1, 1] box (only after the ray origin if it is within the box).  We empirically find that it improves the estimation of ray hit slightly.
We train the model on 32 A100 GPUs for 250 hours (880k iterations).

\subsection{Chamfer distance}\label{supp sec: chamfer}

There are multiple definition of Chamfer distance in the literature (for example, whether to apply the square on the $\ell_2$ norm in the definition below).
We follow the definition used in PointFlow~\citep{yang2019pointflow} and  LION~\citep{vahdat2022lion}, as they are closer to our method:
\begin{equation}
	CD(\cX, \cY) = \frac{1}{|\cX|} \sum_{x \in \cX} \min_{y \in \cY} \| x- y\|^2_2 +  \frac{1}{|\cY|}  \sum_{y \in \cY} \min_{x \in \cX} \| x- y\|^2_2,
\end{equation}
where $\cX$ and $\cY$ are point clouds containing $|\cX|$ and $|\cY|$ points, respectively.
We follow LION's evaluation protocol, which always use $|\cX| = |\cY|$ and the points are scaled to range from $[-0.5, 0.5]$ with axis-aligned box normalization.

\section{More quantitative results}\label{supp sec: more quant}

\subsection{More on 3D Clip}\label{supp sec: clip full}

\Cref{table: clip full} provides full results on 3D CLIP task mentioned in~\Cref{sec: exp clip}.
Even though ULIP-2~\cite{xue2024ulip} has demonstrated better performance and has an improved text corpus, part of its training pipeline is not available. 
Since our goal is to demonstrate feasibility of integrating shape tokens to align with pretrained CLIP embeddings, we choose OpenShape~\cite{liu2024openshape} for its better maintained codebase.

We compare with pretrained OpenShape models provided by the official repository. 
The model is trained to take both color-less and colored point clouds.
When evaluating with color-less point clouds, rgb value is set to $(0.4, 0.4, 0.4)$, following the value used during training. 
In their paper, when only training a model only with color-less point clouds, they achieve lower accuracies (39.6\% top-1 accuracy on Objaverse-LVIS and 83.6\% top-1 accuracy on ModelNet40). 
ULIP~\cite{xue2023ulip} does not take rgb and we reference the numbers in ULIP-2~\cite{xue2024ulip}.
ULIP-2~\cite{xue2024ulip} provides a pretrained model that takes xyz as input, which we evaluate. The numbers on Objaverse-LVIS match those provide in their ablation study.

While we utilize a pretrained and frozen shape tokenizer as our shape encoder, we achieve competitive accuracies when compared to the end-to-end trained PointBERT shape encoder. 
Our performance is also close to ULIP-2, which is an improved pipeline with better text captions.

We notice that our validation accuracy on ModelNet-40 reached 83.3\% after training for 2 days and started decreasing afterwards, while the validation accuracy on Objaverse-LVIS kept increasing during the entire training.
Since ModelNet40 is not part of the training set, this indicates a distribution mismatch between Objaverse-LVIS (potentially due to the quality of the text captions) and ModelNet40, we believe an in-depth analysis of this observation is out of the scope of the paper.

\subsection{Runtime analysis}\label{supp sec: runtime}

In \Cref{table: runtime}, we report the runtime of  (a) computing \Ours from 16,384 input points, (b) sampling 16,384 points from \Ours with 100 Euler steps, (c)  sampling the image-conditioned latent flow matching model with 100 Euler steps, and (d) total time to generate a point cloud containing 16,384 points from a single image.
We measure the runtime with various combinations of hardware (H100, A100) and floating point precision (bfloat16, float32).
Encoding point clouds into \Ours is fast (\eg, 25 \ms on A100 with bfloat16), since it is a feed-forward model.
Sampling point clouds or \Ours requires numerical integration and calling the flow matching models multiple times.
Under the settings, generating a point cloud from a single image takes $>7.5$ seconds using H100 and bfloat16.
There is usually a trade-off between reducing the number of steps, numerical integration method (\eg, first order, second order, \etc), runtime, and generation quality.
Utilizing advancement in diffusion model speedup is future work.

\subsection{Ablation on latent dimension and dataset}
\label{sec: full ablation}

We train multiple shape tokenizers on ShapeNet and Objaverse and with different total latent dimensions. 
\Cref{table: shapenet reconstruction} shows the results of models trained on ShapeNet and evaluated on ShapeNet.
Shape tokenizers achieve better trade-off between the total latent dimension and reconstruction error.

\Cref{table: reconstruction objaverse full} shows models trained on ShapeNet and Objaverse, evaluated on Objaverse and GSO. 
We also create an oracle baseline by comparing with real point clouds sampled independently from the surfaces (shown as Real in \Cref{table: reconstruction objaverse}).

The oracle "real \# points" method in the table is constructed by sampling the real point clouds independently from the reference point cloud.
When the number of points is the same as the reference point cloud (\eg, 8129 and 8192, respectively), it provides the lower-bound on chamfer distance, and when the number of points is smaller than the reference point clouds (\eg, 2048 and 8192, respectively), we randomly sample with replacement to match the number of points in the reference point cloud.

We observe that increasing latent dimension from 8 to 16 improves the results but saturates from 16 to 32. 
This observation is inline with findings by other latent representations (\eg, by~\citet{xiang2024structured}). 
Thus, we choose the $1024 \times 16$ as the main model when integrating shape tokens to various applications.

\section{More visualizations}\label{supp sec: more vis}

This section presents more visualization results and is structured as the following:
\begin{enumerate}
    \item \Cref{supp sec: img-to-3d objaverse} discusses single-image-to-3D results on Objaverse test set (videos \indexhtml);
    \item \Cref{supp sec: img-to-3d gso} talks about single-image-to-3D results on Google Scanned Objects (videos \indexhtml);
	\item \Cref{supp sec: recon_dense_gso} presents reconstruction, densification, and $uvw-xyz$ deformation of point clouds on Google Scanned Objects (videos are \indexhtml);
    \item \Cref{supp sec: room} shows reconstructed point clouds of HM3D houses;
	\item \Cref{supp sec: mult samples} showcases multiple independent samples from the same input image (see videos \indexhtml);
	\item \Cref{supp sec: neural render objaverse} describes neural rendering results  (see videos \indexhtml);
	\item \Cref{supp sec: pcd} provides point cloud filtering results;
    \item \Cref{supp sec: shapenet generations} presents generated point clouds by the unconditional generative models on ShapeNet of various methods.
\end{enumerate}

\subsection{Single-image-to-3D on Objaverse}\label{supp sec: img-to-3d objaverse}

\Suppweb, we present $>100$ videos of single-image-to-3D results on the Objaverse test set.  In the videos, we first show the input image, then the sampled point cloud from the sampled \Ours from the input view. Finally, we rotate the viewpoints.
As can be seen from the videos, our results follow input image closely from the input viewpoint and have plausible 3D structures when seen from other viewpoints.
In the results, the \Ours are sampled with 250 steps using Heun's method, and the point clouds are sampled with 100 steps using Heun's method.  We use classifier free guidance with scale of 5.

\subsection{Single-image-to-3D on  GSO}\label{supp sec: img-to-3d gso}

\begin{figure*}[!th]
	\centering
	\includegraphics[width=0.8\linewidth]{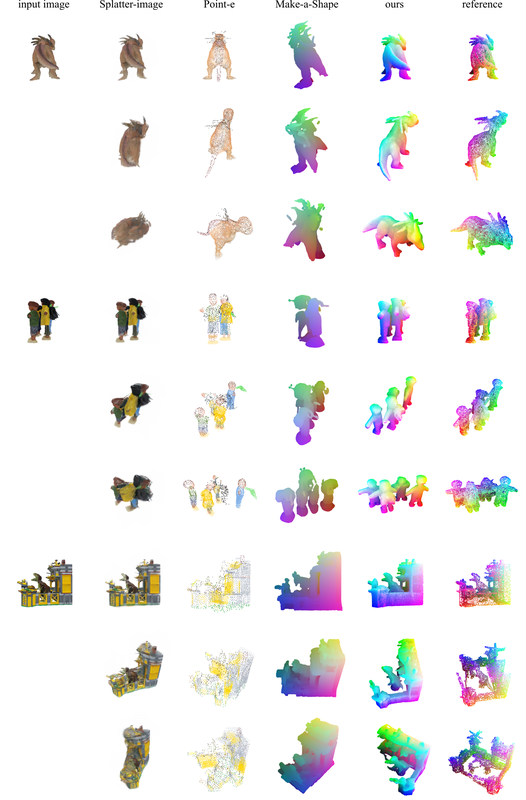}
	\caption{Single-image-to-3D results on Google Scanned Objects (1/3).   Each row block shows different views of the same generated 3D representation from the same input image.}
	\label{figure: gso comparison 1}
\end{figure*}

\begin{figure*}[!th]
	\centering
	\includegraphics[width=0.8\linewidth]{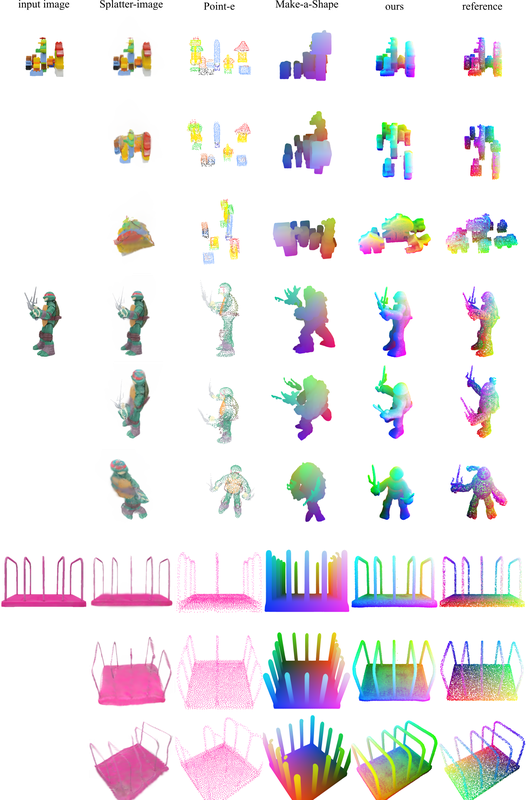}
	\caption{Single-image-to-3D results on Google Scanned Objects (2/3). Each row block shows different views of the same generated 3D representation from the same input image.}
	\label{figure: gso comparison 2}
\end{figure*}

\begin{figure*}[!th]
	\centering
	\includegraphics[width=0.8\linewidth]{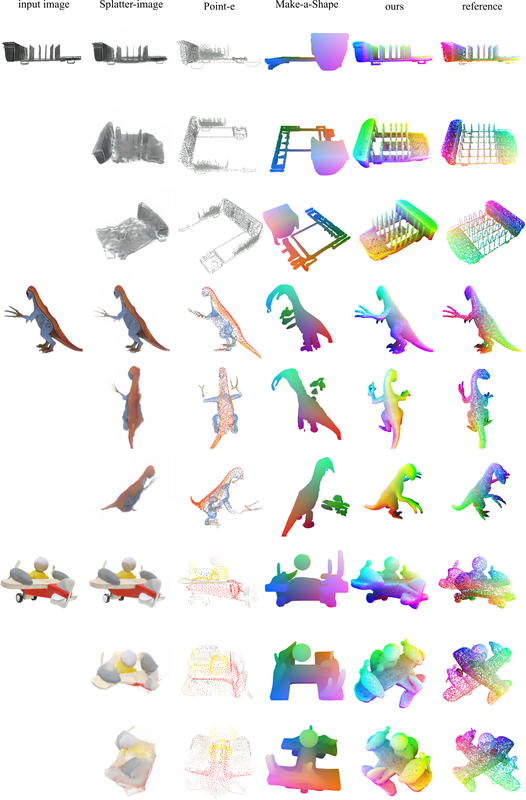}
	\caption{Single-image-to-3D results on Google Scanned Objects (3/3). Each row block shows different views of the same generated 3D representation from the same input image. }
	\label{figure: gso comparison 3}
\end{figure*}

\Suppweb, \Cref{figure: gso comparison 1}, \Cref{figure: gso comparison 2} and \Cref{figure: gso comparison 3},  we present more than 20 single-image-to-3D results on Google Scanned Objects (GSO).
We also present results from recent single-image-to-3D methods using the same input image:
\begin{itemize}[leftmargin=*]
	\item Point-e~\cite{nichol2022point}, which is trained on a proprietary dataset containing several millions meshes.  It first generates 1024 points, then uses another model to upsample to 4096 points. It models the joint distribution of a point set with a fixed number of points and cannot sample arbitrary number of points.
	\item Splatter-image~\cite{szymanowicz2024splatter}, which is a recent method that takes an image as input and predicts 3D Gaussian splats representing the scene. It also models RGB color.
	The model is trained on Objaverse.  Along the same line of works as splatter images, recent methods \cite{xu2024grm, li2023instant3d} often use an additional multiview image diffusion model to generate multiview images from a single image, then apply the multiview images to a model that is similar to Splatter-image to construct 3D Gaussian splats.
	We think Splatter-image reasonably demonstrate the performance of such methodology without using the additional image diffusion model.
	\item Make-a-Shape~\cite{hui2024make}, which is a recent method that represents voxel grids of signed distance functions with packed and pruned wavelet coefficients. A diffusion model is learned to generate the representation given a single image. The model is trained on $>$10 million meshes from 18 datasets, including Objaverse.
\end{itemize}
We present these results for the reader's reference, and we want to emphasize that they are not intended for direct comparison.
The models differ in their training data (\eg, Point-e is trained on a proprietary dataset) and underlying mechanisms (\eg, Splatter-image is not a generative model and our model assumes the input camera parameters are known).
In general, we find it difficult to establish completely fair comparisons of image-to-3D methods.  We hope our code and model release can help improve the situation.
In the results, the \Ours are sampled with 250 steps using Heun's method, and the point clouds are sampled with 100 steps using Heun's method.  We use classifier free guidance with scale equal to 5.

\subsection{Reconstruction and densification on  GSO}\label{supp sec: recon_dense_gso}

\Suppweb,  we present $>$ 20 videos of point clouds sampled from the \Ours computed from input point clouds in Google Scanned Objects (GSO).
The input point clouds contain 16,384 points, and we sample 262,144 points ($16\times$) to demonstrate the densification capability.
We also color the point clouds with their initial coordinate in the noise space (uvw) to demonstrate the deformation trajectory from the noise space (uvw) to the ambient space (xyz).
As can be seen, the trajectories smoothly vary in 3D.

\subsection{Reconstructed HM3D houses}
\label{supp sec: room}

\begin{figure*}[t]
	\centering
	\begin{subfigure}[t]{0.85\linewidth}
		\centering
		\includegraphics[width=\linewidth]{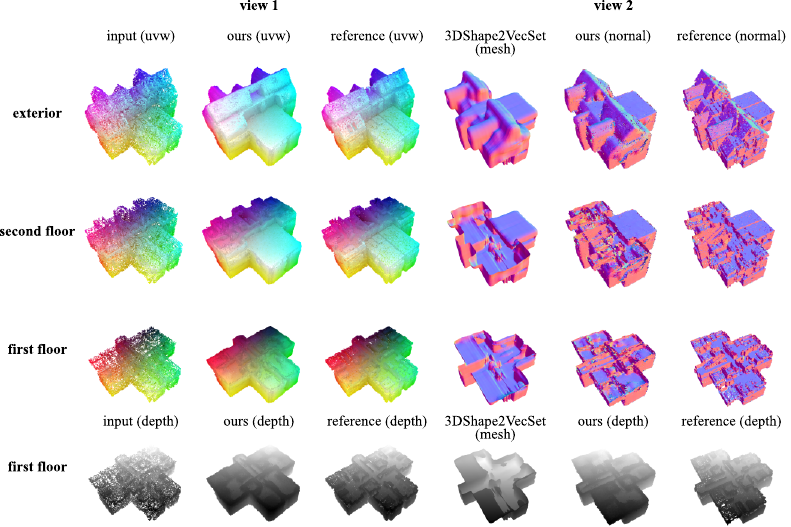}
		\caption{House A}
	\end{subfigure}
	\\[1em]
	\begin{subfigure}[t]{0.85\linewidth}
		\centering
		\includegraphics[width=\linewidth]{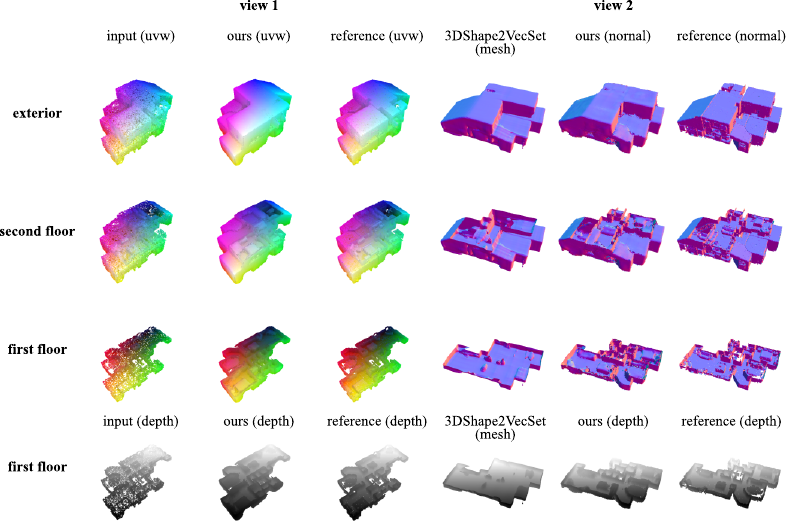}
		\caption{House B}
	\end{subfigure}
	\vspace{-3mm}
	\caption{Reconstruction of HM3D houses~\cite{ramakrishnan2habitat}. Given 16,384 input points of the entire house, we compute \Ours and sample 1 million points. We visualize individual floors by removing points whose z coordinates are larger than a threshold. A reference point cloud containing 100k points is provided for comparison and account for artifacts due to point-cloud visualization (\eg, holes and aliasing).  We also provide meshes reconstructed by 3DShape2VecSet.
    }
	\label{fig: hm3d}
\end{figure*}

\Cref{fig: hm3d} shows two multi-level houses from HM3D~\cite{ramakrishnan2habitat}. 
We use the shape tokenizer trained on Objaverse that takes 16,384 input points sampled from the entire house and outputs $1024$ shape tokens of $16$ dimension. 
Note that the input point cloud is box-normalized to [-1, 1], so furniture like chairs becomes intricate details in the scene. 
In order to show the interior of the house, we manually crop the points for top floors (by thresholding the $z$ coordinate during visualization). 
As can be seen, the reconstructed point clouds not only contain exterior but also the interior structures like walls, sofas, and beds. 
However, we also see blurriness in detailed regions like stair cases.
We also see loss of details in 3DShape2VecSet results, potentially due to trained only on watertight meshes, which are difficult to model interior structures and furniture. 
Extending shape tokenization from object-centric scenes to large scenes like multi-level houses is interesting future work.

\subsection{Multiple samples from same image}\label{supp sec: mult samples}

\begin{figure*}[t]
	\centering
	\includegraphics[width=\linewidth]{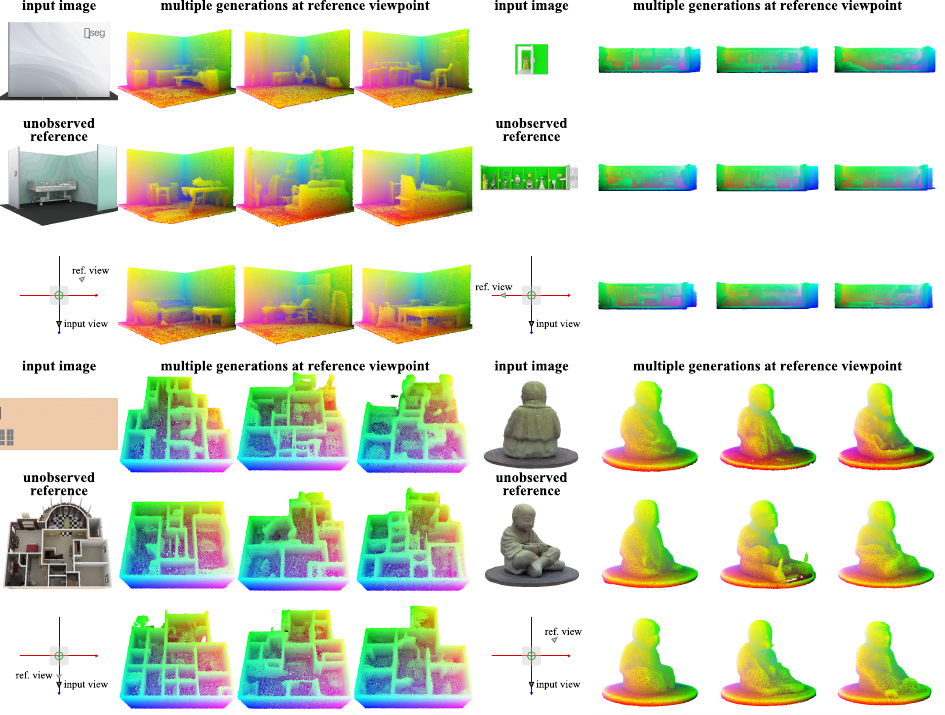}
	\caption{We generated 9 point clouds independently from the same input image (top left image of each set). We provide the rendered image of the meshes at the same viewpoint as a reference (middle left). No human selection was conducted.  Note that the model does not observe the reference images. Mesh credits~\cite{ exhibitbook_qseg_isolation, onironauta_monk_statue, binkley_galactic_truckstop, homedesign3d_new_project}. }
	\label{figure: 3d multigen}
\end{figure*}

In \Cref{figure: 3d multigen} and~\suppweb, we present results of point clouds sampled from independently sampled \Ours from the same input image.  The input images are from Objaverse test set.
As can be seen, the model can generate diverse samples when the input image is ambiguous while matching the input image.
In the results, the \Ours are sampled with 250 steps using Heun's method, and the point clouds are sampled with 100 steps using Heun's method.  We do not use classifier-free guidance in these examples.

\subsection{Neural rendering results on Objaverse}\label{supp sec: neural render objaverse}

\Suppweb, we present results of neural rendered normal maps from input point clouds from Objaverse test set.
We also present results from screened Poisson reconstruction~\cite{kazhdan2013screened} and Pointersect~\cite{chang2023pointersect}.
Screened Poisson reconstruction first reconstructs a mesh from the input point cloud, then renders the normal maps.
Since the input point clouds do not contain vertex normal, we use Open3D to estimate vertex normal by computing principle components of local point clouds.
Screened Poisson reconstruction is sensitive to the quality of the vertex normal.
We use the implementation of screened Poisson reconstruction in Open3D with depth=7, and we remove the vertices with density in the last 5\% percentile. We empirically find the settings produces slightly better results in our experiments.

Pointersect is a neural rendering method that takes a target ray and an input point and estimates the intersection point between the ray and the underlying shape represented by the point cloud.
We find it preserve high frequency details in the rendered normal maps, but it is also sensitive to the input point cloud and thus its results often contain high frequency noise.
Our neural rendering model takes a target ray and \Ours computed from the input point cloud, and it estimates the intersection point between the ray and the underlying shape represented by the point cloud.
The normal estimation is more robust to input-point configurations, however, we also observe smoothing in the rendered normal maps.

\subsection{Point cloud filtering results}\label{supp sec: pcd}

\begin{figure*}[t]
	\centering
	\includegraphics[width=\linewidth]{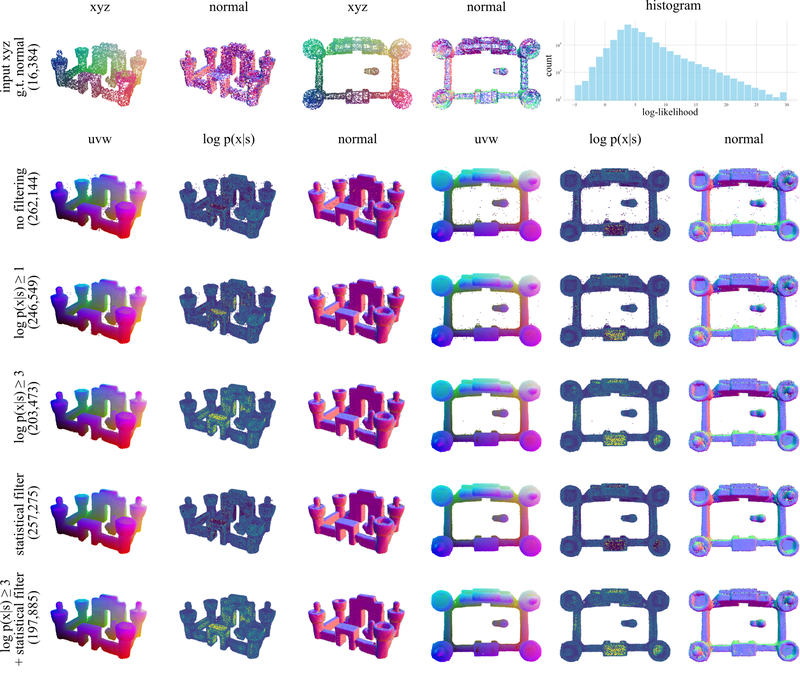}
	\vspace{-7mm}
	\caption{Denoising with log-likelihood.  We sample 262k points from $p(x | \bs)$.  Due to error from the numerical integration, a small number of points contain noise. We compute exact log-likelihood $\log p(x|\bs)$ for each point and use the values to filter. Log-likelihood filtering is complementary to the standard statistical outlier filtering, which also effectively filters noisy points. }
	\label{fig: gso logp 00019}
\end{figure*}

\begin{figure*}[t]
	\centering
	\includegraphics[width=\linewidth]{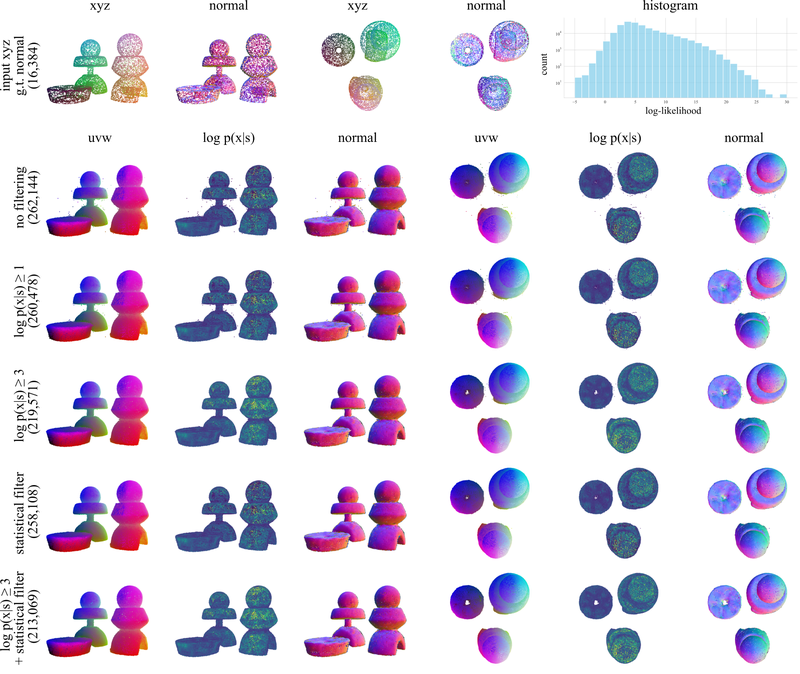}
	\vspace{-7mm}
	\caption{Denoising with log-likelihood.  We sample 262k points from $p(x | \bs)$.  Due to the finite capacity of neural network and the large number of points, a small number of points contain noise. We compute exact log-likelihood $\log p(x|\bs)$ for each point and use the values to filter. Log-likelihood filtering is complementary to the standard statistical outlier filtering, which also effectively filters noisy points. }
	\label{fig: gso logp 00028}
\end{figure*}

In \Cref{fig: gso logp 00019} and \Cref{fig: gso logp 00028}, we show point clouds before and after filtering by the log-likelihood computed using the instantaneous change of variables technique~\cite{chen2019neural}.
In the results of the paper, we sample point clouds containing more than 200 thousands of points using numerical integration of the ordinary differential equations of flow matching with finite number of steps (\eg, 100) of uniform step sizes.
Since we sample individual points in the large number of points independently, a small number of points may contain error from the numerical integration.
As a result, some points may be away from the surfaces after the integration.
We notice that we can calculate the log-likelihood of the sampled points with a small number of steps (\eg, 25) and with Euler method using uniform step sizes and filter the sampled point cloud by thresholding the log-likelihood.
As can be seen from the results in \Cref{fig: gso logp 00028} (\eg, the center hole in the top down view), the filtering is effective and can remove points that are not removed by the standard statistical outlier removal method.
In all results of the paper, we apply log-likelihood filter to remove 10\% of the points with lowest log-likelihood, and then apply statistical outlier removal with neighbor size of 3 and standard deviation to be 2~\cite{Zhou2018}.
We emphasize that the filtering is only conducted for visualization, all quantitative evaluations are conducted on the unfiltered point clouds.

\subsection{Generated results of unconditional LFM on ShapeNet}
\label{supp sec: shapenet generations}

\begin{figure*}
    \centering
    \begin{subfigure}[t]{\linewidth}
        \centering
        \includegraphics[width=\linewidth]{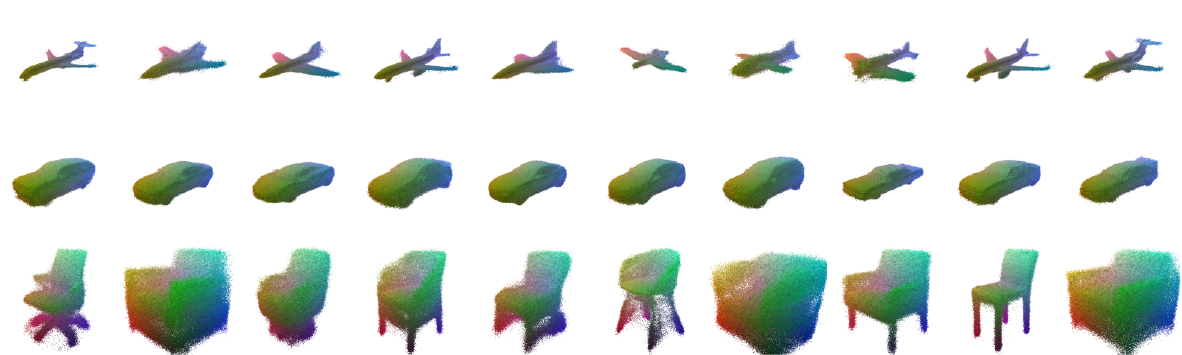}
        \caption{PointFlow}
    \end{subfigure}
    \\[1em]
    \begin{subfigure}[t]{\linewidth}
        \centering
        \includegraphics[width=\linewidth]{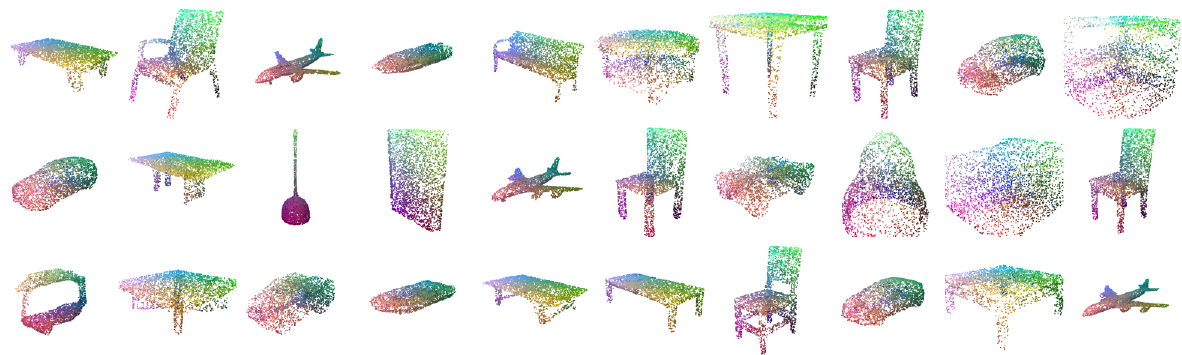}
        \caption{LION}
    \end{subfigure}
    \\[1em]
    \begin{subfigure}[t]{\linewidth}
        \centering
        \includegraphics[width=\linewidth]{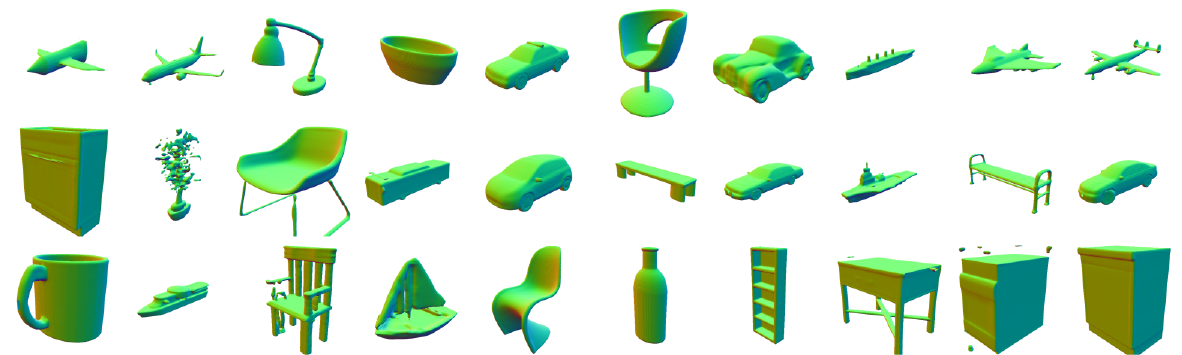}
        \caption{3DShape2VecSet}
    \end{subfigure}
    \caption{Unconditional generation results on ShapeNet dataset (1/2). }
    \label{figure: unconditional shapenet 1}
\end{figure*}

\begin{figure*}
    \centering
    \begin{subfigure}[t]{\linewidth}
        \centering
        \includegraphics[width=\linewidth]{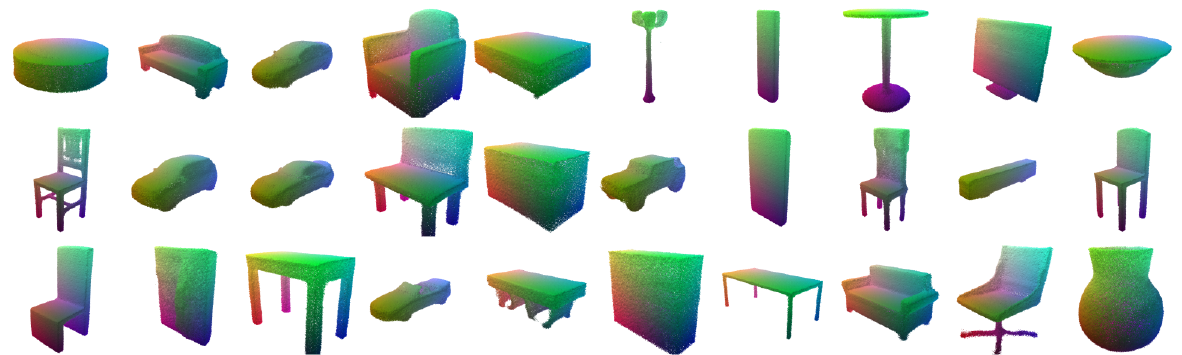}
        \caption{DPF}
    \end{subfigure}
    \\[1em]
    \begin{subfigure}[t]{\linewidth}
        \centering
        \includegraphics[width=\linewidth]{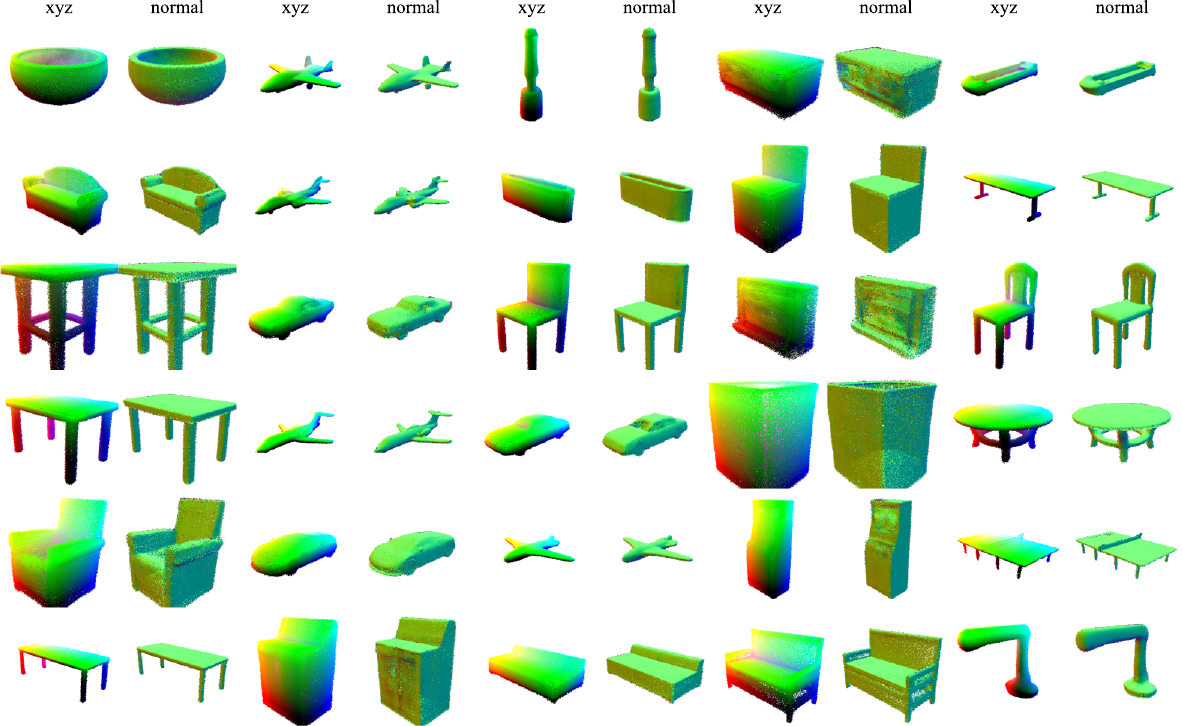}
        \caption{Ours}
    \end{subfigure}
    \caption{Unconditional generation results on ShapeNet dataset (2/2). }
    \label{figure: unconditional shapenet 2}
\end{figure*}

In \Cref{figure: unconditional shapenet 1} and \Cref{figure: unconditional shapenet 2}, we provide generated results from PointFlow, LION, 3DShape2VecSet, DPF, and ours. 
Since PointFlow does not provide the unconditional generative model used in their paper and only provides generative models trained for airplanes, cars, and chairs, we provide samples from these models. 
LION only generates point clouds containing 2048 points, shown by the sparseness in the visualization.
3DShape2VecSet generates occupancy grids and uses marching cube to extract meshes.
However, they train on closed meshes that are often smoother, reflected in their generated samples.
DPF directly learn the point distribution by end-to-end learning in the ambient space (\ie, directly on point cloud), and generates high quality point clouds. 
Our model also generates high quality point clouds and we visualize the zero-shot estimated normal. 
As can be seen, the normal follows the surfaces nicely.

\begin{figure*}[t]
	\centering
	\includegraphics[width=\linewidth]{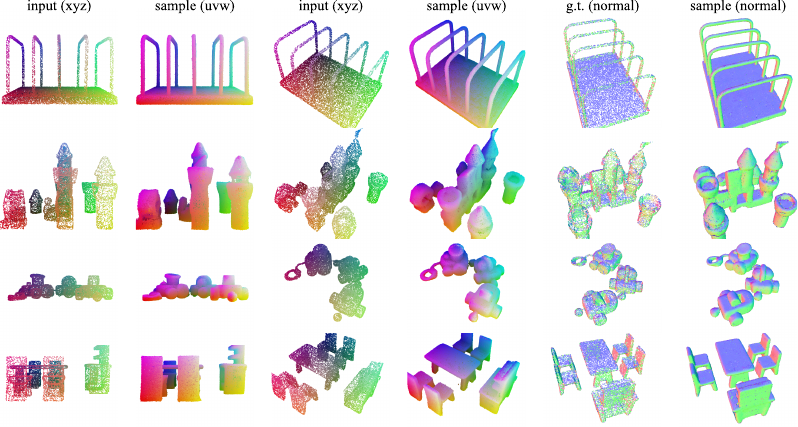}
	\vspace{-7mm}
	\caption{Reconstruction, densification, and normal estimation on GSO dataset.  We tokenize the unseen 16,384 input points (xyz only) with the tokenizer trained on Objaverse, and \iid sample the resulted $p(x|\bs)$ for 262,144 points. We color the input points according to their xyz values and the sampled points according to their initial noise's uvw values and their estimated normal (last column).  Note that we do not provide normal as input to the tokenizer.
	}
	\label{fig: gso reconstruction}
\end{figure*}

%% file: tables/related_works_archive.tex
\newcommand{\gyes}{\textcolor[RGB]{0,140,0}{\ding{51}}}  %
\newcommand{\ono}{\textcolor{orange}{\ding{55}}}
\def\lspace{0.5ex}

\begin{table*}[t]
	\caption{\textbf{Recent latent 3D representations.} The table provides a summary of recent 3D representations and their properties.  We compare the properties that are relevant to machine learning applications.  \textit{Minimal preprocessing} indicates how easy is it to utilize a 3D dataset (\eg, do we need to convert data to watertight meshes, do we need optimization radiance fields to acquire the actual training dataset).  \textit{Continuous latent} indicates whether the 3D representation is fully differentiable (\eg, no graph topology or sparsity patterns).  Total latent dimension indicates the total size to represent one scene.  Note that there may be multiple variants of the same method with different latent dimensions. We choose the representative one in each paper.  * indicates concurrent works.
	}
	\label{table: related works}
	\centering
	\begin{adjustbox}{max width=\linewidth}
		\begin{tabular}{llllccl}
			\toprule
			\textbf{name} & \textbf{what is modeled} & \textbf{minimal preprocessing} & \textbf{continuous latent} & \textbf{total latent dimension} & \textbf{input to encoder} & \textbf{training dataset} \\
			\midrule
			\addlinespace[1ex]
			DDPM-PointCloud~\cite{ddpm_pointcloud} & p(xyz) & \gyes & \gyes & 256 & point cloud & ShapeNet \\
			PointFlow~\cite{yang2019pointflow} & p(xyz) & \gyes & \gyes & 512 & point cloud & ShapeNet \\
			\blue{Ours-shapenet} & p(xyz) & \gyes & \gyes & 32 $\times$ 16 & point cloud & ShapeNet \\
			\blue{Ours-Objaverse} & p(xyz) & \gyes & \gyes & 1024 $\times$ 16 & point cloud & Objaverse \\
			\addlinespace[\lspace]
			\graymidrule
			\addlinespace[\lspace]
			Point-E~\cite{nichol2022point} & fixed size point set & \gyes & \gyes & - & - & proprietary dataset \\
			LION~\cite{lion} & fixed size point set & \gyes & \gyes & 128 + 8192 & point cloud & ShapeNet \\
			\addlinespace[\lspace]
			\graymidrule
			\addlinespace[\lspace]
			3DShape2VecSet~\cite{zhang20233dshape2vecset} & occupancy field & \ono (watertight mesh) & \gyes & 512 $\times$ 32 & point cloud & ShapeNet-watertight \\
			3DILG~\cite{zhang20223dilg} & occupancy field & \ono (watertight mesh) & \ono (discrete and number of cubes) & 512 $\times$ 2 & point cloud & ShapeNet-watertight \\
			Michelangelo~\cite{zhao2023michelangelo} & occupancy field & \ono (watertight mesh) & \gyes & 512 $\times$ 64 + 768 & point cloud & ShapeNet, 3D cartoon monster \\
			CLAY~\cite{zhang2024clay} & occupancy field & \ono (watertight mesh) & \gyes & 2048 $\times$ 64 & point cloud & Objaverse \\
			Dora~\cite{chen2024dora}*  & occupancy field  & \ono (watertight mesh) & \gyes  & 1280 $\times$ 64 & point cloud & Objaverse \\
			Pandora3D~\cite{yang2025pandora3d}*  & occupancy field  & \ono (watertight mesh) & \gyes  & 2048 $\times$ 64 & point cloud & \makecell[tl]{Objaverse, ObjaverseXL, ABO, BuildingNet, \\HSSD, Toy4k, polygone dataset, proprietary} \\

			\addlinespace[\lspace]
			\graymidrule
			\addlinespace[\lspace]
			Direct3D~\cite{wu2024direct3d} & occupancy grid & \ono (watertight mesh) & \gyes & 3 $\times$ 32 $\times$ 32 $\times$ 16 & point cloud & proprietary dataset \\
			XCube~\cite{ren2024xcube} & occupancy grid & \ono (watertight mesh) & \ono (hierarchical) & 16\textsuperscript{3} $\times$ 16 + more & occupancy grid & ShapeNet, Objaverse \\
			\addlinespace[\lspace]
			\graymidrule
			\addlinespace[\lspace]
			Diffusion-SDF~\cite{chou2023diffusion} & SDF field & \ono (watertight mesh) & \gyes & 768 & point cloud & ShapeNet-watertight (Acronym), YCB \\
			MOSAIC-SDF~\cite{yariv2024mosaic} & SDF field & \ono (watertight mesh and optimization) & \gyes & 1024 $\times$ (3+1+7\textsuperscript{3}) & - & ShapeNet-watertight, scalable 3D captioning dataset \\
			TripoSG~\cite{li2025triposg}*  &  SDF field  &  \ono (watertight mesh) &  \gyes  & 4096 $\times$ 64 & point cloud & Objaverse. ObjaverseXL \\
			Hunyuan3D 2.0~\cite{zhao2025hunyuan3d}*  &  SDF field  &  \ono (watertight mesh) &  \gyes  & 3072 $\times$ 64 & point cloud & Objaverse. ObjaverseXL, and more \\

			\addlinespace[\lspace]
			\graymidrule
			\addlinespace[\lspace]
			Make-A-Shape~\cite{hui2024make} & SDF grid & \ono (watertight mesh) & \ono (sparsity pattern) & ~9M & - & 18 datasets \\
			\addlinespace[\lspace]
			\graymidrule
			\addlinespace[\lspace]
			Volume Diffusion~\cite{tang2023volumediffusion} & radiance field & \ono (inference feedforward network) & \gyes & 32\textsuperscript{3} $\times$ 4 & multiview images & Objaverse 
            \\

			\addlinespace[\lspace]
			\graymidrule
			\addlinespace[\lspace]
			TRELLIS~\cite{xiang2024structured}*  & 3D Gaussian  & - (150 multiview DINOv2 features) & \ono (sparsity pattern)  & $\sim$20000 $\times$ 11 & sparse feature grid & Objaverse. ObjaverseXL, ABO, 3D-future, HSSD \\
			\bottomrule
		\end{tabular}
	\end{adjustbox}
\end{table*}

%% file: tables/3d_clip_full.tex
\begin{table*}[t]
	\caption{Zero-shot text classification. The first two row block shows comparison between OpenShape with a PointBERT encoder and OpenShape with frozen \OursABV + MLP adapter.
    The last row block include other current methods for reference.}
	\label{table: clip full}
	\vspace{-2mm}
	\centering
	\begin{adjustbox}{max width=\linewidth}
		\begin{tabular}{lllccccccc}
			\toprule
            &
			\multirow{2}{*}{Pipeline} &
            \multirow{2}{*}{\makecell{Shape\\encoder}} &
            \multirow{2}{*}{\makecell{Input\\xyz}} &
            \multirow{2}{*}{\makecell{Input\\RGB}} &
			\multirow{2}{*}{Training Data} &
			\multicolumn{2}{c}{Objaverse-LVIS}  &
			\multicolumn{2}{c}{ModelNet40}
			\\
			&
            &
			&
            &
            &
            &
			top-1   &
			top-5  &
			top-1   &
			top-5
			\\
			\midrule
			1 & OpenShape & PointBERT &  %
            \cmark & &
           Objaverse~\citep{objaverse}, ShapeNet~\citep{shapenet}, 3D-FUTURE~\citep{fu20213d}, ABO~\citep{collins2022abo} &

			42.6 &
			73.1 &
			\textbf{84.7} &
			\textbf{97.4}
			\\
			2 & OpenShape & ST  &
            \cmark & &
			 Objaverse~\citep{objaverse}, ShapeNet~\citep{shapenet}, 3D-FUTURE~\citep{fu20213d}, ABO~\citep{collins2022abo}  & %
			\textbf{48.4} &
			\textbf{75.5} &
			78.6 &
			93.4
			\\
            \midrule
            3 & OpenShape & PointBERT  &  %
            \cmark & \cmark &
			 Objaverse~\citep{objaverse}, ShapeNet~\citep{shapenet}, 3D-FUTURE~\citep{fu20213d}, ABO~\citep{collins2022abo}  & %
			46.8 &
			77.0 &
			84.4 &
			98.0
			\\
            4 & OpenShape & PointBERT  &  %
            \cmark & \cmark &

			 Objaverse~\citep{objaverse}, ShapeNet~\citep{shapenet} & %
			46.5 &
			76.3 &
			82.6 &
			96.9
			\\
			5 & OpenShape & ST  &
            \cmark & & 
            Objaverse~\citep{objaverse}, ShapeNet~\citep{shapenet} &

			47.9  &  %
			75.1  &   %
			80.6  &  %
			94.6  %
			\\
			\midrule
			6 & ULIP & PointBERT &  %
            \cmark & & 

			 Objaverse~\citep{objaverse}, ShapeNet~\citep{shapenet} & %
			34.9 &
			61.0 &
			69.6 &
			85.9
			\\
			7 & ULIP-2 & PointBERT & %
            \cmark  &  &
			 Objaverse~\citep{objaverse}, ShapeNet~\citep{shapenet} &  %
			48.9 &
			77.1 &
			84.1 &
			97.3
			\\
			\bottomrule
		\end{tabular}
	\end{adjustbox}
\end{table*}

%% file: tables/recon_shapenet_full.tex
\begin{table*}[!t]
	\caption{Reconstruction error on ShapeNet. We provide the full tables used to plot \Cref{fig: architecture}. Chamfer distances are computed using 2048 points.}
	\label{table: shapenet reconstruction}
	\vspace{-2mm}
	\centering
	\begin{adjustbox}{max width=\linewidth}
		\begin{tabular}{lccccc|cc}
			\toprule
			&
			PointFlow &  %
			{\OursABV}  &  %
			{\OursABV}  &  %
			{\OursABV}  &  %
			{\OursABV}  &  %
			{3DShape2VecSet}  & %
			LION    %
			\\
			\midrule
			latent  &
			$512$ &
			$512 \ {(32 {\times} 16)}$ &
			$2048 \ {(32 {\times} 64)}$ &
			${4096 \ (256 {\times} 16)}$ &
			$8192 \ {(512 {\times} 16)} $ &
			${4096 \ (512 {\times} 8)} $   & %
			$8320$   %
			\\
			CD ($\times 10^{-3}$) $\downarrow$ &
			$1.75 \pm 1.53$ &   %
			${0.82} \pm 1.19$ &  %
			$0.75 \pm 0.53$   & %
			{$0.68 \pm 0.60$}  &  %
			${0.65} \pm 1.3$  & %
			{$0.79 \pm 3.6$}  & %
			$0.84 \pm 2.4$  %
			\\
			\bottomrule
		\end{tabular}
	\end{adjustbox}
\end{table*}

%% file: tables/recon_objaverse_full.tex
\begin{table*}[!tb]
	\caption{Reconstruction error on Objaverse and GSO datasets. We provide the full table including our ablation study on \OursABV with different total latent dimensions and trained on ShapeNet or Objaverse datasets. We compute Chamfer distances with 2048 and 8192 points, and they are of unit $10^{-4}$.  The first row block are models trained on ShapeNet, \OursABV in the second block are trained on Objaverse, and TRELLIS trained on ObjaverseXL and three other datasets. * indicates normal estimated by Open3D.}
	\label{table: reconstruction objaverse full}
	\vspace{-2mm}
	\centering

    \begin{adjustbox}{max width=\linewidth}
		\begin{tabular}{lllcccccc}
			\toprule
			&
			\multirow{2}{*}{\makecell{training\\ data}}&
			\multirow{2}[4]{*}{\makecell{latent}} &
			\multicolumn{3}{c}{Objaverse}   &
			\multicolumn{3}{c}{Google scanned objects}
			\\
			&
			&
			&
			CD@2048$\downarrow$ &
			CD@8192 $\downarrow$ &  %
			normal ($^\circ$)$\downarrow$&
			CD@2048$\downarrow$ &
			CD@8192 $\downarrow$ &
			Normal  ($^\circ$) $\downarrow$
			\\
			\midrule

			{Michelangelo} &
			{shapenet} &
			{$512{\times}64$}&
			{${32.2} \pm 47.4$} &
			{${27.5} \pm 47.7$} &
			{${28.7} \pm 13.7$} &
			{${15.6} \pm 18.3$} &
			{${10.8} \pm 17.4$} &
			{${18.3} \pm 11.3$}
			\\

			{ST} &
			{shapenet} &
			{$32{\times}16$}&
			{$12.3 \pm 7.3$} &
			{$7.5 \pm 5.0$} &
			{$44.4 \pm 9.2$} &
			{$10.9 \pm 5.0$} &
			{$5.6 \pm 3.7$} &
			{$34.0 \pm 13.3$}
			\\
			{ST} &
			{shapenet} &
			{$32{\times}64$}&
			{$11.0 \pm 6.4$} &
			{$6.3 \pm 4.2$} &
			{$40.9 \pm 10.5$} &
			{$10.1 \pm 4.6$} &
			{$4.9 \pm 3.2$} &
			{$31.4 \pm 13.7$}
			\\
			{3DShape2VecSet} &
			{shapenet} &
			{$512{\times}8$}&
			{${10.7} \pm 15.8$} &
			{${6.5} \pm 13.6$} &
			{$\mathbf{20.1} \pm 12.0$} &
			{${8.6} \pm 11.9$} &
			{${4.1} \pm 11.0$} &
			{$\mathbf{12.2} \pm 8.0$}
			\\
			{ST} &
			{shapenet} &
			{$256{\times}16$}&
			{$7.9 \pm 4.8$} &
			{$3.7 \pm 2.1$} &
			{$38.3 \pm 10.3$} &
			{$8.2 \pm 3.3$} &
			{$3.2 \pm 1.6$} &
			{$29.0 \pm 12.6$}

			\\
			{ST} &
			{shapenet} &
			{$512{\times}16$}&
			{$\mathbf{6.8} \pm 4.5$} &
			{$\mathbf{2.8} \pm 1.7$} &
			{${32.1} \pm 11.2$} &
			{$\mathbf{7.4} \pm 3.0$} &
			{$\mathbf{2.6} \pm 1.2$} &
			{${23.2} \pm 12.0$}
			\\
			\midrule
			{TRELLIS} &
			{objaverseXL+ and 3 others} &
			{${\sim}20000{\times}11$}&
			{$5.6 \pm 3.5$} &
			{$2.0 \pm 1.3$} &
			{$\mathbf{15.4} \pm 10.9$} &
			{$6.7 \pm 2.8$} &
			{$2.2 \pm 0.90$} &
			{$\mathbf{10.0} \pm 4.92$}
			\\
			{ST} &
			{objaverse} &
			{$1024{\times}8$}&
			$5.6 \pm 4.0$ &
			$1.8 \pm 1.2$ &
			$22.5 \pm 10.2$ &
			$6.7 \pm 3.0$ &
			$2.0 \pm 0.85$ &
			$15.1 \pm 8.26$
			\\
			{ST} &
			{objaverse} &
			{$1024{\times}16$}&
			$\mathbf{5.4} \pm 4.0$ &
			$\mathbf{1.6} \pm 1.1$ &
			$19.0 \pm 8.84$ &
			$6.6 \pm 2.9$ &
			$1.9 \pm 0.83$ &
			$13.9 \pm 6.84$
			\\
			{ST} &
			{objaverse} &
			{$1024{\times}32$}&
			{$5.5 \pm 4.1$} &
			{$1.7 \pm 1.3$} &
			{$18.2 \pm 8.2$} &
			{$\mathbf{6.6} \pm 2.9$} &
			{$\mathbf{1.9} \pm 0.82$} &
			{${13.7} \pm 5.43$}
			\\
			\midrule
			\multicolumn{3}{l}{real 2048 points}  &
			$5.1 \pm 3.9$ &
			$3.2 \pm 2.5$ &
			$25.3 \pm 11.1$* &
			$6.3 \pm 2.9$ &
			$4.0 \pm 1.8$ &
			$17.6 \pm 8.74$*
			\\
			\multicolumn{3}{l}{real 8192 points}  &
			-  & %
			$1.3 \pm 1.1$ &
			$20.5 \pm 9.95$* &
			-  &  %
			$1.6 \pm 0.77$ &
			$14.3 \pm 7.68$*
			\\
			\bottomrule
		\end{tabular}
	\end{adjustbox}

	\vspace{-2mm}

\end{table*}